%% file: main.tex
\documentclass[lettersize,journal]{IEEEtran}
\usepackage{amsmath,amsfonts}
\usepackage{algorithmic}
\usepackage{array}
\usepackage[caption=false,font=normalsize,labelfont=sf,textfont=sf]{subfig}
\usepackage{textcomp}
\usepackage{stfloats}
\usepackage{url}
\usepackage{verbatim}
\usepackage{graphicx}
\usepackage{cite}
\usepackage{multirow}
\usepackage{multicol}
\usepackage{bm}
\usepackage{booktabs}
\usepackage{amsmath,epsfig}
\usepackage{dsfont}

\usepackage{bbding}
\usepackage{amssymb}
\usepackage{pifont}
\newcommand{\cmark}{\ding{51}}%
\newcommand{\xmark}{\ding{55}}%

\usepackage{xcolor}

\usepackage[ruled,linesnumbered]{algorithm2e}

\definecolor{visualorange}{HTML}{F4B183}
\definecolor{textgreen}{HTML}{70AD47}
\definecolor{applegreen}{rgb}{0.55, 0.71, 0.0}
\definecolor{iccvblue}{rgb}{0.21,0.49,0.74}

\usepackage{makecell}
\usepackage{fancybox}
\usepackage[many]{tcolorbox}
\newcommand{\revision}[1][\textcolor{black}]{#1}

\begin{document}

\title{Hyper-modal Imputation Diffusion Embedding with Dual-Distillation for Federated Multimodal Knowledge Graph Completion}

\author{Ying Zhang, \IEEEmembership{Member,~IEEE},
Yu Zhao, Xuhui Sui, Baohang Zhou, Xiangrui Cai, 
Li Shen, \IEEEmembership{Member,~IEEE},
\\Xiaojie Yuan, Dacheng Tao, \textit{Fellow, IEEE}
\thanks{This research was supported by the National Natural Science Foundation of China (No. 62272250, 62572260), the Natural Science Foundation of Tianjin, China (No. 22JCJQJC00150), and the Fundamental Research Funds for the Central Universities, Nankai University (No. 63253232). The work of Yu Zhao was supported by the China Scholarship Council (No. 202506200065). (\textit{Corresponding author: Yu Zhao}.)}
\thanks{Ying Zhang, Yu Zhao, Xuhui Sui, Xiangrui Cai, Xiaojie Yuan are with the College of Computer Science, VCIP, DISSec, Nankai University, Tianjin, China (e-mail: yingzhang@nankai.edu.cn, zhaoyu@dbis.nankai.edu.cn, suixuhui@dbis.nankai.edu.cn, caixr@nankai.edu.cn, yuanxj@nankai.edu.cn).}
\thanks{Baohang Zhou is with the School of Software, Tiangong University, Tianjin, China (e-mail: zhoubaohang@tiangong.edu.cn).}
\thanks{Li Shen is with the Shenzhen Campus of Sun Yat-sen University, Shenzhen, China (e-mail: mathshenli@gmail.com).}
\thanks{Dacheng Tao is with the College of Computing and Data Science, Nanyang Technological University, Singapore (e-mail: dacheng.tao@ntu.edu.sg).}
\thanks{\copyright 2026 IEEE. Personal use of this material is permitted.  Permission from IEEE must be obtained for all other uses, in any current or future media, including reprinting/republishing this material for advertising or promotional purposes, creating new collective works, for resale or redistribution to servers or lists, or reuse of any copyrighted component of this work in other works.}
\thanks{Published in IEEE Transactions on Multimedia.
DOI: 10.1109/TMM.2026.3705201}
}

\maketitle

\input{scripts/abstract}

\input{scripts/intro}
\input{scripts/related_work}

\input{scripts/method}

\input{scripts/experiments}
\input{scripts/conclusion}

\bibliographystyle{IEEEtran}
\bibliography{mybib}

\input{photos/authors}
\vfill

\newpage
\appendices

\input{scripts/appendix}

\end{document}

%% file: scripts/abstract.tex
\begin{abstract}

With the increasing multimodal knowledge privatization requirements, multimodal knowledge graphs in different institutes are usually decentralized, lacking of effective collaboration system with both stronger reasoning ability and transmission safety guarantees.
In this paper, we propose the Federated Multimodal Knowledge Graph Completion (FedMKGC) task, aiming at training over federated MKGs for better predicting the missing links in clients without sharing sensitive knowledge.
We propose a framework named MMFeD3-HidE for addressing multimodal uncertain unavailability and multimodal client heterogeneity challenges of FedMKGC. 
(1) Inside the clients, our proposed Hyper-modal Imputation Diffusion Embedding model (HidE) recovers the complete multimodal distributions from incomplete entity embeddings constrained by available modalities.
(2) Among clients, our proposed Multimodal FeDerated Dual Distillation (MMFeD3) transfers knowledge mutually between clients and the server with logit and feature distillation to improve both global convergence and semantic consistency. 
We propose a FedMKGC benchmark for a comprehensive evaluation, consisting of a general FedMKGC backbone named MMFedE, datasets with heterogeneous multimodal information, and three groups of constructed baselines.
Experiments conducted on our benchmark validate the effectiveness, semantic consistency, and convergence robustness of MMFeD3-HidE.

\end{abstract}


%% file: scripts/intro.tex
\section{Introduction}
Multimodal knowledge graphs (MKGs) \cite{zhu2022mmkgsurvey,liang2024mmkgsurvey} organize graph structures composed of relational triples \textit{(head entity, relation, tail entity)} and their visual and textual attributes as Figure \ref{fig:intro}, which have been widely in multimodal knowledge-intensive tasks \cite{marino2019okvqa,ding2022mukea,sun2020mmkgrs,TMM_KG_Object_Detection}.
Due to incomplete construction and new knowledge emergence, Multimodal Knowledge Graph Completion (MKGC) task \cite{xie2017IKRL,xie2016DKRL} has been widely-explored \cite{wang2021rsme,zhao2022mose,chen2022mkgformer,shang2024lafa} to reason the missing links in MKGs with multimodal information.
For example, in Figure \ref{fig:intro}, given a query \textit{(Kobe Bryant, Team member, ?)}, MKGC aims to predict the tail entity \textit{L.A. Lakers} with graph structures, entity images, and descriptions.

\input{figs/intro}

In real world, the MKGs are usually decentralized in different institutes due to commercial interests or data regulations, such as open-sourced MKGs DBpedia \cite{auer2007dbpedia}, Wikidata \cite{vrandevcic2014wikidata}, Freebase \cite{bollacker2008freebase}, and domain-specific MKGs in e-commerce \cite{wang2024amazonKG,xu2021alimemmkg-e-commerce} or medical health \cite{rotmensch2017EHRKG}. 
To address the data-isolation problem,
Federated Learning \cite{yang2019FMLsurvey,mcmahan2017FedAVG,kairouz2021advancesFLsurvey} (FL) has been proposed for enabling cooperatively training a strong global model without data transmission.
However, the federated learning on multimodal knowledge graphs, particularly concerning privacy-sensitive graph structures, entity images, and entity descriptions, remains an underexplored area. 
Existing Federated KGC approaches \cite{chen2021FedE, chen2022FedEC, zhu2023FedLU,meng2024fedean} mainly focus on modeling graph structures in unimodal KGs, while MKGC approaches \cite{wang2021rsme,zhao2022mose,li2023imf,shang2024lafa} all focus on centralized MKGs with full data availability, leaving the collaboration of federated MKGs unexplored.
Despite the great progress of existing federated multimodal learning studies \cite{feng2023fedmultimodal,yu2023CreamFL,xiong2022unifiedMMFL}, they cannot address the graph structure modality and cross-silo knowledge challenges in federated institutes.

In this paper, we propose the \textit{Federated Multimodal Knowledge Graph Completion (FedMKGC)} task, aiming at training a global model to complete the missing links in all client MKGs for better overall MKGC performance while preventing decentralized multimodal information transmission. 
Specifically, as shown in Figure \ref{fig:intro}, there is a set of client MKGs and a server, where the structural, visual, and textual modalities in client MKGs preserve different aspects of knowledge for the same entity \textit{Kobe Bryant}. 
The federated learning over cross-institute MKGs could securely obtain a more comprehensive representation for entity \textit{Kobe Bryant} to improve the local reasoning. 
The federated learning on MKGs provides great potential in the real world for addressing increasing multimodal knowledge privatization and secure sharing requirements.

It is non-trivial to propose address the FedMKGC task due to its following novel challenges:

\textbf{\textit{(1) Multimodal uncertain unavailability without reconstruction supervision.}}
In the real world, exclusive knowledge, such as intimate relations, sensitive images, and personal descriptions, can be fragmented in client MKGs and thus unavailable to other MKGs. 
Incomplete MKG construction \cite{zhang2023maco,zhang2024unleashing} also leads to modality unavailability.
Thus, the federated MKGs face \textit{uncertain missing modalities problem, i.e., one or two of the textual or visual modalities are randomly unavailable}. which results in inconsistent multimodal semantics in client MKGs and hinders the reasoning.
Existing incomplete multimodal learning approaches \cite{zhao2021MMIN, yuan2021transformerreconstruction,zeng2022TagTMM,huan2023unimf,wang2024diffusedEmoRec} usually reconstruct the multimodal representations for further application. However, they cannot be directly applied to FedMKGC since they usually require modality-complete samples for training, while the missing modalities of certain entities are inherently unavailable for both training and inference.
It poses a lack of reconstruction supervision challenge of uncertain missing modalities for FedMKGC.

\textbf{\textit{(2) Multimodal client heterogeneity with inconsistent semantics.} 
}
The knowledge semantics in client MKGs are usually non-identically distributed. 
For example, in Figure \ref{fig:intro}, though three client MKGs all have entity \textit{Kobe Bryant}, they store different aspects of his knowledge from varying neighbors, descriptions, and images. 
In terms of structural modality, client MKGs have different relational schemas \cite{chen2021FedE}, leading to their heterogeneous topologies and structures. 
In terms of visual and textual modalities, their semantics are also diverse between clients due to different knowledge coverage and concentrations of institutes. 
The multimodal uncertain unavailability mentioned above also increases client heterogeneity.
The multimodal heterogeneity poses a challenge for robust global convergence \cite{li2020FedProx,li2022federatednoniid,TMM24FL_NonIID_MMdata} for FedMKGC.

To address above challenges, we propose a novel FedMKGC framework \textbf{MMFeD3-HidE}, \textbf{H}yper-modal \textbf{i}mputation \textbf{d}iffusion \textbf{E}mbedding with \textbf{M}ulti-\textbf{M}odal \textbf{FeD}erated \textbf{D}ual \textbf{D}istillation. 
Firstly, inspired by missing data imputation research \cite{yoon2018gain,luo2018nipsGRUI}, we formulate the incomplete multimodal entity embeddings as hyper-modal data vectors consisting of all modalities and having inherent missing data points.
This way, the available modalities could lead the missing ones to learn from the same reconstruction distributions to maintain semantic consistency and multimodal integrity.
We exploit the diffusion model to capture the distribution of incomplete hyper-modalities and recover the complete ones from additional Gaussian noises stepwise, and propose to maximize masked variational bound to provide supervision from available modalities.
Secondly, we propose MMFeD3 to mutually transfer knowledge between clients and the server through logit and feature distillation.
The feature distillation further improves semantic consistency by bringing the imputed entity embeddings in the client model and the incomplete ones in the server model closer, while the logit distillation improves the convergence robustness through mutual enhancement in the decision level.

To thoroughly assess the FedMKGC performance, we propose the FedMKGC benchmark, consisting of federated MKG datasets, basic backbone MMFedE, and three groups of baselines. 
Extensive experiments conducted on our benchmark validate the advantages of MMFeD3-HidE over other baselines, including effectiveness, semantic consistency, convergence stability, and efficiency.
The contributions of our paper can be summarized as follows:

\begin{itemize}
    \item To the best of our knowledge, we are the first to study the FedMKGC task and address its uncertain missing modalities and multimodal client heterogeneity challenges. 
    \item We propose a novel framework MMFeD3-HidE with hyper-modal imputation diffusion embedding for missing modality reconstruction, and federated logit-feature dual-distillation for addressing above challenges.
    \item We construct the FedMKGC benchmark with datasets, backbone, and baselines. Experiments demonstrate the effectiveness, semantic consistency, and convergence efficiency of MMFeD3-HidE.
\end{itemize}

The paper is organized as follows: Section \ref{sec:related_work} introduces related works on FedMKGC, Section \ref{sec:benchmark} demonstrates the FedMKGC task formulation and benchmark construction, Section \ref{sec:method} presents our proposed method to address the two challenges, Section \ref{sec:experiments} describes the experiments for validation, and Section \ref{sec:conclusion} presents the conclusion of the paper.

%% file: figs/intro.tex
\begin{figure}[t]
     \centering
         \includegraphics[width=0.45\textwidth]{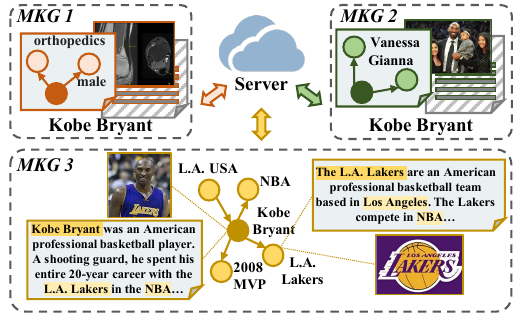}
             \vspace{-10pt}
         \caption{Toy example of the FedMKGC task for decentralized MKGs. The client MKGs have uncertain unavailable modalities (\textit{slash shadowed}) and heterogeneous multimodal semantics.}
    \vspace{-18pt}
\label{fig:intro}
\end{figure}

%% file: scripts/related_work.tex
\section{Related Work} \label{sec:related_work}
\input{tables/FedMMbenchmarks}

\textbf{Incomplete Multimodal Learning (MML).}
Incomplete MML studies reconstruct the 
Generative-based approaches impute the missing modalities with GAN-based \cite{goodfellow2020GAN,shang2017vigan}, AE-based \cite{vincent2008DAE,baldi2012autoencoders,tran2017missingcra}, flow-based \cite{wang2023distributionICCV} networks to generate a single missing modality.
Reconstruct-based approaches obtain complete modalities with AE-based \cite{zhao2021MMIN}, GNN-based \cite{lian2023gcnet}, Transformer-based \cite{yuan2021transformerreconstruction,zeng2022mmin-ensemble,zeng2022TagTMM,huan2023unimf}, or Diffusion-based \cite{wang2024diffusedEmoRec} reconstruction networks.
Distillation-based approaches \cite{li2024correlationKD,wei2023mmanet} distill full-modal knowledge of teacher model to student model with missing modalities. 
They all require ground-truths of the missing modalities to train the model and transfer to the test phase, leaving supervision of inherent missing modalities in both training and testing phases in MKG entities untackled.

\textbf{Federated Multimodal Learning.}
Federated MML studies aim to jointly train a model with decentralized multimodal clients, which has great improvements in diverse tasks including vision-language understanding \cite{yu2023CreamFL,liu2020FLVLgrounding,guo2023pfedpromptFLVL}, Health \cite{dai2024FLbraintumor,borazjani2024MMFLTMI}, IoT \cite{zhao2022multimodalIoT,TMM24FL_NonIID_MMdata}, Human Activity Recognition \cite{xiong2022unifiedMMFL}.
FedMultimodal benchmark \cite{feng2023fedmultimodal} includes tasks such as Emotion Recognition (ER) \cite{poria2018MELDEmotionRecognition,cao2014cremaDEmotionRecognition}, Multimodal Action Recognition (MAR) \cite{soomro2012ucf101ActionRecognition,monfort2019MiTActionRecognition}, Social Media (SM) \cite{kiela2020hatefulmemesSocialMedia,alam2018crisismmdSocialMedia}, as shown in Table \ref{tab: MMFLdatasets}.
Existing studies neglect graph structure modality that is locally learned and distinguished from other modalities, and cross-silo setting \cite{kairouz2021advancesFLsurvey,do2023ICCVcross-siloFL} that has fewer client counts and more local data than cross-device setting. 
\revision{
Moreover, Federated MML studies also investigate the real-world noises, such as data corruption \cite{fang2023ICCVcorruption}, label corruption \cite{feng2023fedmultimodal}, and missing modalities \cite{xiong2023crossmodel,chen2022fedmsplit,dai2024FLbraintumor,che2024R3_FML_incomplete,wang2024R3fedmmr}, in which graph structural modality and inherent missing modalities in both training and testing problems are not explored. 
}



\textbf{Multimodal Knowledge Graph Completion. }
Multimodal knowledge graph completion (MKGC) \cite{xie2016DKRL,xie2017IKRL} aims to complete the missing links in MKGs with multimodal information. 
They \cite{wang2021rsme,zhao2022mose,li2023imf,chen2022mkgformer,zhao2024CMR,shang2024lafa,zhang2024mygo} extend the unimodal knowledge graph embeddings \cite{bordes2013TransE,trouillon2016complex,sun2018RotatE} with the textual and visual embeddings under the assumption of complete modalities. 
For the occasionally missing modality in the datasets, they usually pad them with zero or random values. 
MACO \cite{zhang2023maco} and AdaMF \cite{zhang2024unleashing} highlight the missing multimodal information in MKGC and propose GAN-based architectures to generate the missing visual or text modality. 
However, they focus on the single-modal missing situation and cannot handle uncertain missing modalities. 
In our paper, we focus on the uncertain missing modalities problem, where one or two of the visual and textual modalities could be randomly missing. 

Moreover, existing MKGC methods \cite{wang2021rsme,zhao2022mose,li2023imf,chen2022mkgformer,zhao2024CMR,shang2024lafa} all focus on centralized settings where all entities, relations, triples, and visual/textual modalities are available and unified. 
To achieve jointly training of cross-institute MKGs with untransmittable fragmented knowledge, our paper proposes to study the FedMKGC task, where relational triples, entity images, and descriptions are decentralized and jointly trained through federated dual-distillation. 

\textbf{Federated Knowledge Graph Completion. }
\revision{
Federated Knowledge Graph Completion is a sub-domain of Federated Graph Learning (FGL) studies \cite{2024TNNLSsurveyFLGNN,zhang2021FGLsurvey}, which aims to collaboratively train graph mining models over decentralized graph data while preserving data privacy. 
Federated KGC studies focus on federated learning over several multi-relational graph structures by preventing relation embeddings from sharing and learning global structural entity embeddings \cite{chen2021FedE}. }
FedE \cite{chen2021FedE} proposes to extend FedAvg \cite{mcmahan2017FedAVG} and average aggregate entity embeddings, accomplished with permutation matrices to map the server entity embeddings to the client ones. 
FedEC \cite{chen2022FedEC} further augment FedE with a contrastive learning objective to constrain the embedding updates.
FedLU \cite{zhu2023FedLU} proposes logit distillation between server and clients to address the structural heterogeneity problems.
\revision{
Some FedKGC methods focus on other problems such as peer-to-peer setting \cite{peng2021FKGE}, unlearning \cite{zhu2023FedLU,liu2024R3FedKGEUnlearningDiffusion}, cross-domain \cite{huang2022fedcke}, global relation embeddings \cite{zhang2022FedR}, diffusion structural entity embedding \cite{zhao2025R3dfedkg}, entity codebooks \cite{zhang2025MMFedMKGC}, named entity recognition \cite{chen2025R3federatedMNER}, attacks \cite{yang2024BlackBoxAttackFedKGE,zhou2024poisoningAttackFedKGE} and defends \cite{hu2023quantifydefendFedKGE}, IoT Services \cite{sun2025TITFedKGE}, and Neighbor Prediction \cite{liu2025NPFedKGC}.}

In our paper, we focus on the FedMKGC task to focus on multimodal federated learning including structural, visual, and textual modalities. 
The multimodal uncertain unavailability and multimodal client heterogeneity problems in FedMKGC are also non-explored by FedKGC studies. 

%% file: tables/FedMMbenchmarks.tex
\begin{table}[!b]
\vspace{-20pt}

    \begin{center}
            \caption{Overview of Federated Multimodal Learning Datasets.}    \label{tab: MMFLdatasets}

    \resizebox{0.48\textwidth}{!}{
    \small
\begin{tabular}{clccr}
\toprule
\textbf{Task}&\textbf{Dataset} & \textbf{FL setting} & \textbf{Modalities} & \textbf{Total Instance} \\
\midrule
\multirow{2}{*}{\textbf{ER}} & MELD&cross-device & Audio,Text & 9,718\\
&CREMA-D&cross-device & Audio,Video&4,798\\
\midrule
\multirow{3}{*}{\textbf{MAR}}& UCF101 & cross-device&Audio, Video&6,837\\
&MiT10&cross-device&Audio, Video&41.6K\\
& MiT51 & cross-device& Audio, Video&157.6K \\
\midrule
\multirow{2}{*}{\textbf{SM}} &Hateful-Memes & cross-device & Image, Text & 10.0K \\
&CrisisMMD & cross-device & Image, Text & 18.1K\\
\midrule
\textbf{MKGC} & \textbf{FedMKGC} & \textbf{cross-silo} & \makecell[c]{\textbf{Structure,}\\\textbf{Image, Text}} & \makecell[r]{\textbf{Entity:14,541}\\\textbf{Triple: 310,116}} \\
\bottomrule

    \end{tabular}}
    \end{center}

    \vspace{-10pt}
    \end{table}

%% file: scripts/method.tex

\section{FedMKGC Task and Benchmark} \label{sec:benchmark}



\subsection{Federated MKGC Task Formulation} 
The federated multimodal knowledge graph completion (FedMKGC) task focuses on conducting MKGC on all client MKGs without data sharing, where there is a set of client MKGs $\mathcal{C}=\{\mathcal{G}^c\}$ and a central server. 
An MKG is denoted as $\mathcal{G}^c=\{\mathcal{E}^c,\mathcal{R}^c,\mathcal{T}^c,\mathcal{V}^c,\mathcal{D}^c\}$, with entity set $\mathcal{E}^c$, relation set $\mathcal{R}^c$, 
triple set $\mathcal{T}^c$ that denotes the relation between entities $\{(h,r,t)\}\subseteq \mathcal{E}^c\times\mathcal{R}^c\times\mathcal{E}^c$, 
entity image set $\mathcal{V}^c$, and entity description set $\mathcal{D}^c$. 
In each client $\mathcal{G}^c$, given a query $q=(h,r,?)$ with head entity $h$ and relation $r$, 
the model aims to rank all the entities in client MKG $e \in \mathcal{E}^c$ with a score function $f(h,r,t) \to \mathbb{R}^{|\mathcal{E}^c|}$ and make the true tail entity $t$ the highest score. 
The head prediction is formed as tail prediction $(t, r^{-1}, h)$ \cite{bordes2013TransE} for unified modeling.
The client multimodal features ${E}_m^c,m\in\{v,d\}$ are extracted from fixed pretrained encoders BERT \cite{devlin2019bert} and ViT \cite{dosovitskiy2020ViT}.
Main notations used in this paper are shown in Appendix A.



\subsection{Non-IID MKG Partition}
Since there is no federated MKG dataset, we partition the existing MKG to create authentic non-IID distributions similar to real-world MKGs. 
(1) \textit{MKG triple partition} with relation IDs \cite{feng2023fedmultimodal,chen2021FedE} that clients have no overlapping relation types.
We construct the FedMKGC benchmark based on the FB15K-237 dataset \cite{bordes2013TransE}, which has sufficient multimodal information and is widely adopted \cite{zhao2022mose,chen2022mkgformer,li2023imf}.
The common partition scheme is to partition data through unique client identifier \cite{feng2023fedmultimodal,kairouz2021advancesFLsurvey} to form non-IID distributions, such as participant IDs.
Migrating to the MKGs, the relation IDs are commonly used for partitioning triples between clients \cite{chen2021FedE,zhu2023FedLU}.
Thus, we follow them to partition triples in clients based on relation IDs.
This way, the client KGs concentrate on different relation types, ensuring that the relational schemas and topologies in the client MKGs are heterogeneous.

(2) \textit{MKG multimodal information partition} following Dirichlet distribution \cite{feng2023fedmultimodal,hsu2019dirichletNonIID} that one entity in different clients has different descriptions and images.
Unlike triples, the images and descriptions of an entity do not have client identifiers, thus we need to synthesize the non-IID visual and textual partitions.
We first split the paragraph of entity descriptions into multiple descriptive sentences and crawl multiple images for each entity. 
Then we label the sentences and images with different IDs and partition them following Dirichlet distribution \cite{hsu2019dirichletNonIID,feng2023fedmultimodal,yu2023CreamFL} with $\alpha_{Dir}=0.1$ for high heterogeneity.
After partition, clients have diverse descriptions and images for the same entity.

\input{algs/alg}


\subsection{Missing Modalities Construction}
We randomly generate the modality availability mask of all the entities in $\mathcal{E}^c$ for each client following Bernoulli distribution \cite{chen2022fedmsplit,feng2023fedmultimodal} with an availability rate of $r$.
We exploit the multimodal feature mask $M^c_m \in \mathbb{R}^{|\mathcal{E}^c|\times d_m}$ to represent availability of modality $m$ in client $c$, where $|\mathcal{E}^c|$ is the number of entities in client $c$ and $d_m$ is feature dimension of modal $m$. 
That is, the $i$-th row $M^c_{m,(i,\cdot)}=\mathbf{1}$ if the $i$-th entity has available modality $m$, otherwise $M^c_{m,(i,\cdot)}=\mathbf{0}$.
We assign different masks for visual and text modality and for different clients. 
The features of unavailable entity images or descriptions are randomly padded \cite{zhao2022mose,chen2022mkgformer,zhang2024unleashing}. 

\subsection{Federated MKGC Backbone}\label{sec:MMFedE}
We propose a general backbone named MMFedE with basic average aggregation idea \cite{mcmahan2017FedAVG,chen2021FedE}.
MMFedE aggregates hybrid structural entity embeddings and visual/textual projection weights. 
The structural entity embeddings are locally learned and do not expose privacy \cite{chen2021FedE}, while extracted visual and textual features from pretrained models are preserved in clients for communication efficiency and data protection.
The pipeline of MMFedE is shown in Algorithm \ref{alg: pipeline}. 
\input{figs/method}

\textbf{Client end. }
Clients are trained on local MKGC objectives. 
We denote the fusion method as $\Phi(\cdot)$, and the fused multimodal entity embeddings can be denoted as
$\mathbf{E}^c =\Phi(\mathbf{S}^c,\mathbf{V}^c,\mathbf{D}^c)$ where $\mathbf{V}^c=\mathbf{W}^{c}_v {E}^c_v$
and $\mathbf{D}^c= \mathbf{W}^{c}_d {E}^c_d$ are mapped visual and textual embeddings.
The probability of target entity $t$ among the negative entities is as Equation \eqref{eq: prob_c}:
\begin{equation} \label{eq: prob_c}
\small
\begin{aligned}
p^c(t|(h,r)_i) & = \text{softmax}(f^c(h,r,e)), e\in\{t, N_i\},\\
&= \text{softmax}(f^c(h,r,e; \mathbf{E}^c,\mathbf{R}^c)),
\end{aligned}
\end{equation}
where $\mathbf{R}^c$ is the relation embeddings of client $c$, $N_i$ is the negative entity set of $i$-th triple.
The score function $f(\cdot)$ could be any KG embedding (KGE) decoder \cite{bordes2013TransE,trouillon2016complex,sun2018RotatE}. 
The KGC objective is cross-entropy loss as Equation \eqref{eq: loss_kgc_client}. 
At test time, clients reason with Equation \eqref{eq: prob_c} that $\hat{e}=\arg\max p^c(e|(h,r)_i;\mathbf{E}^c,\mathbf{R}^c),e\in\mathcal{E}^c$ as predicted entity. 
\begin{equation} \label{eq: loss_kgc_client}
\small
\mathcal{L}^c_{KGC}=\sum_{(h,r,t)_i\in\mathcal{G}^c}-\log \left[ p^c(t|(h,r)_i) \right]
\end{equation}

\textbf{Server end. }
At each communication round $ro$, the server first samples a client subset $\mathcal{C}_{ro}$ to collaborate. 
MMFedE distributes both global structural entity embeddings $\mathbf{S}_{ro}^s$ and multimodal feature projection weights $\mathbf{W}_{m,{ro}}, m\in\{v,d\}$.
After the local training, the clients update the local structural embeddings $\mathbf{S}^{c}_{ro+1}$ and multimodal weights $\mathbf{W}_{m,{ro}}^c$ to the server. 
MMFedE hybridly aggregates the structural embeddings with entity existence ratio $\mathds{1}\oslash\sum_{\mathcal{C}_{ro}} \mathbf{v}^c$ as weights \cite{chen2021FedE}, and 
the multimodal projection matrices with triple ratio $\alpha^c=\frac{|\mathcal{T}^c|}{\sum_c |\mathcal{T}^c|}$ as weights \cite{mcmahan2017FedAVG}.


\section{Method} \label{sec:method}
The framework of MMFeD3-HidE is shown in Figure \ref{fig:method}.
The algorithm illustration is in App. B.A. 
We first reconstruct the incomplete multimodal entity embeddings with hyper-modal diffusion imputation, optimized with $\mathcal{L}_{DI}$.
Then we optimize the federated client MKGs with MMFeD3, including logit distillation $\mathcal{L}_{LD}$ and feature distillation $\mathcal{L}_{FD}$. 

\subsection{Hyper-modal Construction}
Inspired by missing data imputation research \cite{yoon2018gain,luo2018nipsGRUI}, we treat multimodal embeddings of an entity as a hyper-modal data vector consisting of all modalities and have some inherent missing data points. 


We define the hyper-modality feature as the concatenation of the multimodal embeddings in structural vector space as $\mathbf{H}^c = [\mathbf{S}^c||\mathbf{V}^c||\mathbf{D}^c]$.
Since the structural modal embeddings are learned based on the structures inside the client MKG, $\mathbf{S}^c$ is the only fully available modality.
Thus, we employed the pre-trained FedE structural embeddings to initialize $\mathbf{S}^c$ for guiding the hyper-modal imputation.
\subsection{Imputation Diffusion Model Processes} \label{sec: diffusion process}
Diffusion models \cite{sohl2015DPM, ho2020DDPM,luo2022understandingDM} typically consist of two processes, i.e., forward process and reverse process.
Given the input incomplete hyper-modal feature $\mathbf{x}_0=\mathbf{H}^c$ as the original state\footnote{The processes of the diffusion model for each client are the same and we omit the superscript $^c$ of $\mathbf{x}_t$ in Section \ref{sec: diffusion process} and \ref{sec: diffusion optim} for convenience. }, the forward process corrupts  $\mathbf{x}_0$ and constructs a series of latent variables $\mathbf{x}_{1:T}$ in the Markov chain by adding Gaussian noises gradually in $T$ steps. 
In the reverse process, the reconstruction network learns to recover the $\hat{\mathbf{x}}_{t-1}$ given $\mathbf{x}_t$, which starts from $\mathbf{x}_T$ and finally recovers the $\hat{\mathbf{x}}_0$ and the output $\hat{\mathbf{x}}_0$ is the recovered complete hyper-modal feature. 
In the end, we impute the $\hat{\mathbf{x}}_0$ to the missing points of $\mathbf{x}_0$ as imputed hyper-modal features $\tilde{\mathbf{x}}$. 

\textbf{{Forward process. }} 
We add the Gaussian noise to $\mathbf{x}_0$ as a Markov chain to form $\mathbf{x}_{1:T}$ according to a variance schedule $\beta_1, ..., \beta_T$, where $t\in \{1, ..., T\}$ denotes diffusion step and $\beta_t$ denotes variance at step $t$. 
The forward process is denoted as Equation \eqref{eq: q_x_t_x_t-1}, $\mathcal{N}$ denotes Gaussian distribution.
\begin{equation} \label{eq: q_x_t_x_t-1}
\small
    q(\mathbf{x}_t|\mathbf{x}_{t-1})= \mathcal{N}(\mathbf{x}_t; \sqrt{1-\beta_t}\mathbf{x}_{t-1}, \beta_t \mathbf{I})
\end{equation}
The forward process can be reparameterized to sample $\mathbf{x}_t$ at an arbitrary time step $t$ \cite{ho2020DDPM} as Equation \eqref{eq:q_x_t_x_0}, where $\alpha_t=1-\beta_t$ and $\bar{\alpha_t}=\Pi_{t'=1}^t\alpha_{t'}$.  
\begin{equation} \label{eq:q_x_t_x_0}
\small
    q(\mathbf{x}_t|\mathbf{x}_0)=\mathcal{N}(\mathbf{x}_t; \sqrt{\bar{\alpha_t}} \mathbf{x}_0, (1-\bar{\alpha_t})\mathbf{I})
\end{equation}

Thus we can further reparameterize the latent variable in each step as $\mathbf{x}_t(\mathbf{x}_0, \boldsymbol{\epsilon}) = \sqrt{\bar{\alpha_t}}\mathbf{x}_0 + \sqrt{1-\bar{\alpha_t}} \boldsymbol{\epsilon} $, where $\boldsymbol{\epsilon} \sim \mathcal{N}(\mathbf{0}, \mathbf{I})$.
In implementation, we ignore the trainable $\beta_t$ by reparameterization and fix $\beta_t$ as constants for simplicity \cite{ho2020DDPM}, linearly increasing from $\beta_1=\beta_{low}$ to $\beta_T = \beta_{up}$ where $\beta_{low}$ and $\beta_{up}$ are hyper-parameters.

\textbf{{Reverse process. }}
In the reverse process, we recover the complete hyper-modal features starting from $\mathbf{x}_T$, by utilizing the reconstruction network to approximate the latent variable states as Equation \eqref{eq:reverse_p_theta}, where $\boldsymbol{\mu}_\theta(\mathbf{x}_t, t)$ and $\mathbf{\Sigma}_\theta(\mathbf{x}_t,t)$ is the Gaussian mean and variance parameters outputted by the reconstruction networks with learnable parameter $\theta$. We employ Cascade Residual Autoencoder (CRA) \cite{tran2017missingcra,zhao2021MMIN} as the reconstruction network $\theta$ for its superior modality reconstruction ability. 
We simply take the mean of the distribution $\hat{\mathbf{x}}_{t-1} = \boldsymbol{\mu}_\theta({\mathbf{x}}_t,t)$ as the predicted variable at previous step. 
Finally, the predicted $\hat{\mathbf{x}}_0$ are seen as the recovered complete hyper-modal features.
\begin{equation} \label{eq:reverse_p_theta}
\small
    p_\theta(\mathbf{x}_{t-1}|\mathbf{x}_t)=\mathcal{N}(\mathbf{x}_{t-1}; \boldsymbol{\mu}_\theta(\mathbf{x}_t, t), \mathbf{\Sigma}_\theta(\mathbf{x}_t,t))
\end{equation}

\textbf{{Imputation process. }}
To fully utilize the existing modalities, we conduct the imputation process to interpolate the generated $\hat{\mathbf{x}}_0$ into the missing slots of the original hyper-modal matrix $\mathbf{x}_0=\mathbf{H}^c$ to obtain the imputed hyper-modal $\tilde{\mathbf{x}}_0$ as Equation \eqref{eq: imputation}, where the $M^c$ guarantees the preservation of the available modalities and $\odot$ is the element-wise product. 
We construct the multimodal mask of client $c$ as ${M}^c = [M^c_s||M^c_v||M^c_d]$, where the feature positions of missing modalities are $0$ and those of available modalities are $1$.
The imputed $\hat{\mathbf{H}}^c$ can be seen as the concatenation of three modal embeddings, where $\hat{\mathbf{V}}^c$ and $\hat{\mathbf{D}}^c$ are imputed complete visual and textual entity embeddings.
Then we could construct the complete multimodal entity embeddings as 
$\hat{\mathbf{E}}^c =\Phi(\mathbf{S}^c,\hat{\mathbf{V}}^c, \hat{\mathbf{D}}^c)$.
\begin{equation} \label{eq: imputation}
\small
\begin{aligned}
    \hat{\mathbf{H}}^c = \tilde{\mathbf{x}}_0& = {M}^c\odot\mathbf{x}_0^c  + (1-{M}^c) \odot \hat{\mathbf{x}}_0^c \\&= {M}^c\odot\mathbf{H}^c  + (1-{M}^c) \odot \hat{\mathbf{x}}_0^c \\
        &= [\mathbf{S}^c||\hat{\mathbf{V}}^c||\hat{\mathbf{D}}^c]
\end{aligned}
\end{equation}


\input{figs/method_diff}
\subsection{Diffusion Imputation Optimization} \label{sec: diffusion optim}

Diffusion models are typically optimized by maximizing the evidence lower bound (ELBO) of the likelihood of the original state $\mathbf{x}_0$ \cite{luo2022understandingDM}.
Based on DDPM \cite{ho2020DDPM}, ELBO can be simplified as $\mathbb{E}_{t,\mathbf{x}_0,\boldsymbol{\epsilon}}\left[\parallel\boldsymbol{\epsilon} - \boldsymbol{\epsilon}_{\theta}(\mathbf{x}_t,t)\parallel_2^2\right]$, where $1\leq t\leq T$. 
\revision{DDPM's reconstruction network outputs the additional noise $\boldsymbol{\epsilon}_{\theta}$ and optimizes the noise to approximate the distribution step by step.
In our paper, we aim to predict the complete hyper-modal distributions conditioned on the input observed modalities; thus, we predict the complete hyper-modal $\mathbf{x}_0$ at each step instead of $\boldsymbol{\epsilon}_{\theta}$. 
Thus, in our HidE, the reconstruction network $\theta$ is optimized to regulate its output $\hat{\mathbf{x}}_\theta(\mathbf{x}_t, t)$ closer to $\mathbf{x}_0$.
This way, the output of the diffusion model can be directly constrained by the observed modalities as strong supervision signals and optimized to remain semantically consistent with the observed modalities. }

However, there are both missing and observed data points in the diffusion processes; we need to design the objective with reconstruction supervision signals that are not affected by the missing modal points. 
Since the diffusion model preserves the feature size and location in both forward and backward processes, it offers a convenient solution for diffusion imputation optimization.
We split the data at each step $\mathbf{x}_t$ to observed data and missing data $\mathbf{x}_t=\{\mathbf{x}_t^{obs},\mathbf{x}_t^{mis}\}$, that $\mathbf{x}_t^{obs}=M^c \odot \mathbf{x}_t$ and $\mathbf{x}_t^{mis}=(1-M^c) \odot \mathbf{x}_t,1\leq t\leq T$.
We propose to constrain the diffusion of available modalities $\mathbf{x}_t^{obs}$ to be similar to their ground-truth features.
\revision{This way, the output of the diffusion model is sampled from the same underlying conditional distribution $p(\mathbf{x}^{complete}|\mathbf{x}^{obs}_0)$, which is the complete hypermodal distribution conditioned on the observed modalities. 
This way, the data points of the missing modalities $\mathbf{x}_t^{mis}$ are also sampled from the conditional distribution $p(\mathbf{x}^{complete}|\mathbf{x}^{obs}_0)$, 
ensuring that the reconstructed hyper-modal embeddings maintain distributional consistency with the original MKGs with missing modalities.}
We elaborate on simplifying ELBO to masked imputation in App. B.B. 
\begin{equation} \label{eq: L_t_imputation}
\small
\begin{aligned}
\mathcal{L}_t& =\mathbb{E}_{q(\mathbf{x}_{t}|\mathbf{x}_0)} \left[ \parallel \hat{\mathbf{x}}^{obs}_\theta(\mathbf{x}_t, t) - \mathbf{x}^{obs}_0 \parallel_2^2  \right]\\
&= \mathbb{E}_{q(\mathbf{x}_{t}|\mathbf{x}_0)} \left[ \parallel M^c\odot\hat{\mathbf{x}}_\theta(\mathbf{x}_t, t) - M^c\odot\mathbf{x}_0 \parallel_2^2  \right],
\end{aligned}
\end{equation}

Since after the optimization, the observed modalities and their reconstructed features are the same $\hat{\mathbf{x}}^{obs}_0=\mathbf{x}^{obs}_0$, 
and the reconstructed hyper-modal $\hat{\mathbf{x}}_0 \sim p_\theta(\mathbf{x}|\hat{\mathbf{x}}^{obs}_0=\mathbf{x}^{obs}_0)$, 
thus estimated missing features $\hat{\mathbf{x}}^{mis}_0$ are also from the same distributions $\hat{\mathbf{x}}_0^{mis} \sim p_\theta(\mathbf{x}^{mis}|\hat{\mathbf{x}}^{obs}_0=\mathbf{x}^{obs}_0)$, which made sure the completed three modalities are from the same distributions and maintain the semantic consistency.

In implementation, we uniformly sample $t\sim \mathcal{U}(1, T)$. The client diffusion imputation loss is as Equation \eqref{eq: Loss_diffusion_imputation}.
\begin{equation} \label{eq: Loss_diffusion_imputation}
\small
    \mathcal{L}_{DI}^c = \mathbb{E}_{t\sim\mathcal{U}(1,T)} \mathcal{L}^c_t
\end{equation}

\subsection{Federated Dual-Distillation Optimization}
MMFeD3 aims to transfer knowledge between the server and clients to improve global convergence robustness, semantic consistency, and inference accuracy, consisting of feature distillation objective and logit distillation objective. 






\textbf{Federated KGC loss. }
In MMFeD3, we jointly train the global parameters with server KGC loss with the client ones for the global parameters to better generalize to all client KGs \cite{liang2020LGFedAvg}.
Server trains the global parameters with KGC loss $\mathcal{L}_{KGC}^{s,c}$ similarly to Equation \eqref{eq: loss_kgc_client} with $p^{s,c}(t|(h,r)_i)$ as Equation \eqref{eq: score_func_c_s}, where global entity embeddings $\mathbf{E}^{s,c}=\Phi(\mathbf{S}^s,\mathbf{V}^{s,c},\mathbf{D}^{s,c})$, where
$\mathbf{V}^{s,c}=\mathbf{W}_v{E}_v^c,~ \mathbf{D}^{s,c}=\mathbf{W}_d{E}_d^c$ are obtained with the global projection weights $\mathbf{W}_m$ and the local multimodal features ${E}_m^c$. The relation embedding $\mathbf{R}^c$ is from local client without transmission \cite{chen2021FedE}.
\begin{equation} \label{eq: score_func_c_s}
\small
    p^{s,c}(t|(h,r)_i) = \text{softmax}(f(h,r,t; \mathbf{E}^{s,c};\mathbf{R}^c))
\end{equation}



\textbf{{Federated logit distillation loss. }}
We co-distillate the knowledge of existing modalities from the server and the knowledge of imputed modalities from clients with logit distillation \cite{hinton2015knowledgedistillation} of KL divergence as Equation \eqref{eq: loss_kd}. The co-distillation could guide the server and clients to converge to the best global and local embeddings. The overall cross-modal distillation loss is denoted as $\mathcal{L}_{LD}^c= \mathcal{L}_{c2s}^c + \mathcal{L}_{s2c}^c$.
\begin{equation} \label{eq: loss_kd}
\small
\left\{
\begin{aligned}
\mathcal{L}_{c2s}^c&=\sum_{(h,r,t)_i\in\mathcal{G}^c} D_\text{KL}(p^c(t|(h,r)_i), p^{s,c}(t|(h,r)_i))\\
\mathcal{L}_{s2c}^c&=\sum_{(h,r,t)_i\in\mathcal{G}^c} D_\text{KL}(p^{s,c}(t|(h,r)_i), p^c(t|(h,r)_i))
\end{aligned}
\right.
\end{equation}

\textbf{{Federated feature distillation loss}}
The federated feature distillation objective aims to improve the expressiveness of global entity embeddings, while also guiding the hyper-modal imputation diffusion network with knowledge from global embeddings aggregated from all clients. As Equation \eqref{eq: feature_distillation_objective}, we minimize the L2 distance between the imputed full-modal client entity embeddings and available-modal server ones.
\begin{equation}\label{eq: feature_distillation_objective}
\small
\mathcal{L}_{FD}^c =\parallel\hat{\mathbf{E}}^c - \mathbf{E}^{s,c}\parallel_2^2
\end{equation}

\subsection{Overall objective }
The overall objective of client $c$ is as Equation \eqref{eq:overall objective}, 
where $\lambda$ controls $\mathcal{L}_{DI}$ in HidE, $\mu$ and $\eta$ control $\mathcal{L}_{LD},\mathcal{L}_{FD}$ in MMFeD3. 
We optimize all the client networks with respective $\mathcal{L}^c$. We keep diffusion model parameters $\theta^c$ local. The server aggregates structural embeddings and multimodal projection weights similar to MMFedE. 
\revision{The local KGC loss $\mathcal{L}_{KGC}^c$ is essential for learning client-specific representations, while the server KGC loss $\mathcal{L}_{KGC}^{s,c}$ ensures the global model's generalization. 
The diffusion imputation loss $\mathcal{L}_{DI}^c$ supervises the multimodal entity embedding reconstruction with available modalities.
The distillation losses ($\mathcal{L}_{LD}, \mathcal{L}_{FD}$) bridge the semantic gap between heterogeneous local and global spaces. }

\begin{equation}\label{eq:overall objective}
\small
\mathcal{L}^c = \mathcal{L}^c_{KGC} + \mathcal{L}^{s,c}_{KGC} + \lambda  \mathcal{L}^c_{DI} + \mu \mathcal{L}_{LD}^c + \eta \mathcal{L}_{FD}^c
\end{equation}
At test time, $\hat{e}=\arg\max p^c(e|(h,r)_i;\hat{\mathbf{E}}^c,\mathbf{R}^c),e\in\mathcal{E}^c$ at each client are predicted as the target entity for a query triple $(h,r)_i$.

%% file: algs/alg.tex
\begin{algorithm}[!t]     
\small
    \SetKwInOut{Input}{Input}
    \SetKwInOut{Output}{Output}
    \SetKwInOut{Require}{Require}
    \Input{Client set $\mathcal{C}=\{\mathcal{G}^c\}$, permutation mapping matrix $\mathbf{P}$, existence vector $\mathbf{v}$, communication rounds $R$}
    Server initialize $\mathbf{S}_0^s$, $\mathbf{W}_{v,0}$, $\mathbf{W}_{d,0}$;\\
    \For{$ro$ in $R$}{
    Sample a client subset $\mathcal{C}_{ro}\subseteq\mathcal{C}$;\\
    \For{$\mathcal{G}^c \in \mathcal{C}_{ro}$}{
    Server distributes $\mathbf{S}_{ro}^c=\mathbf{P}^c\mathbf{S}_{ro}^s$, $\mathbf{W}_{v,ro}$, $\mathbf{W}_{d,ro}$;\\
    $\mathbf{S}^c_{ro+1}, \mathbf{W}^{c}_{v,ro+1}, \mathbf{W}^{c}_{d,ro+1}\leftarrow \arg \min \mathcal{L}_{KGC}^c$;\\
    Client uploads $\mathbf{S}^c_{ro+1}, \mathbf{W}^{c}_{v,ro+1}, \mathbf{W}^{c}_{d,ro+1}$;
    }
    $\mathbf{S}_{ro+1}^s=(\mathds{1}\oslash\sum_{\mathcal{C}_{ro}} \mathbf{v}^c)\otimes \sum_{\mathcal{C}_{ro}} {\mathbf{P}^{c}}^\top \mathbf{S}_{ro+1}^{c} $;\\
    \For{$m \in \{v,d\}$}{
        $\mathbf{W}_{m,ro+1} = \sum_{\mathcal{C}_{ro}} \alpha^{c} \mathbf{W}_{m,ro+1}^{c}$;
    }
    }
        \caption{MMFedE Backbone}
    \label{alg: pipeline}
\end{algorithm}

%% file: figs/method.tex
\begin{figure*}[t]
     \centering
         \centering
         \includegraphics[width=0.99\textwidth]{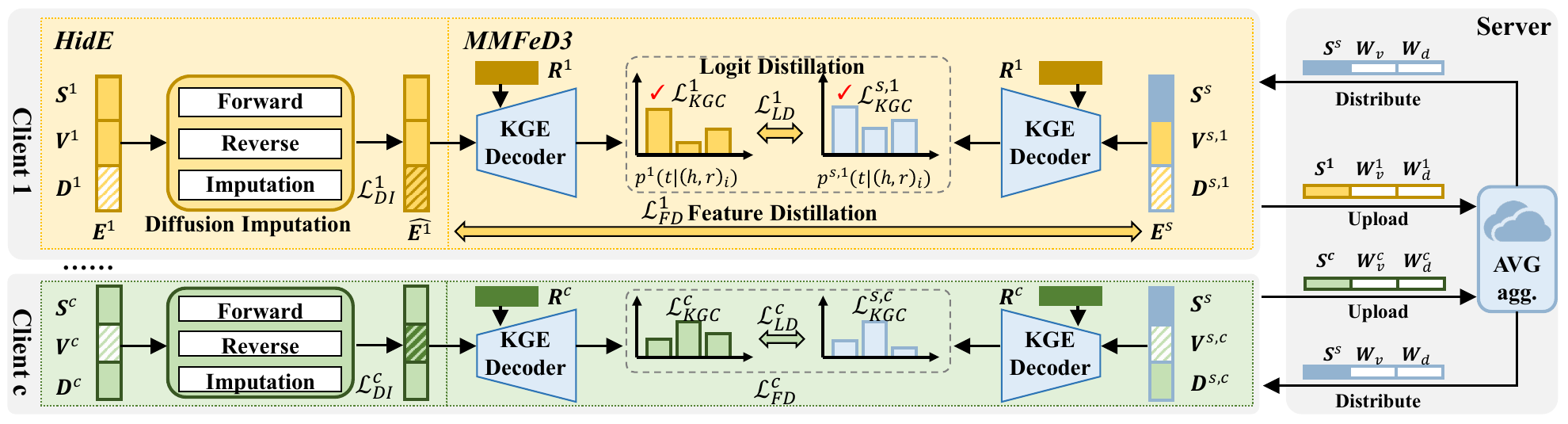}
         \vspace{-10pt}
         \caption{The overview of our MMFeD3-HidE.
         $\mathbf{S}^c,\mathbf{V}^c,\mathbf{D}^c$ represent structural, visual, and textual modalities in client MKGs, where shadowed blocks represent missing.
         The HidE imputes the missing modalities of entities with diffusion imputation.
         The MMFeD3 optimizes the federated MKGC with dual distillation objectives and KGC objectives of clients and server.
         }
         \label{fig:method}
         \vspace{-15pt}
         
\end{figure*}

%% file: figs/method_diff.tex
\begin{figure}[t]
     \centering
         \centering         \includegraphics[width=0.48\textwidth]{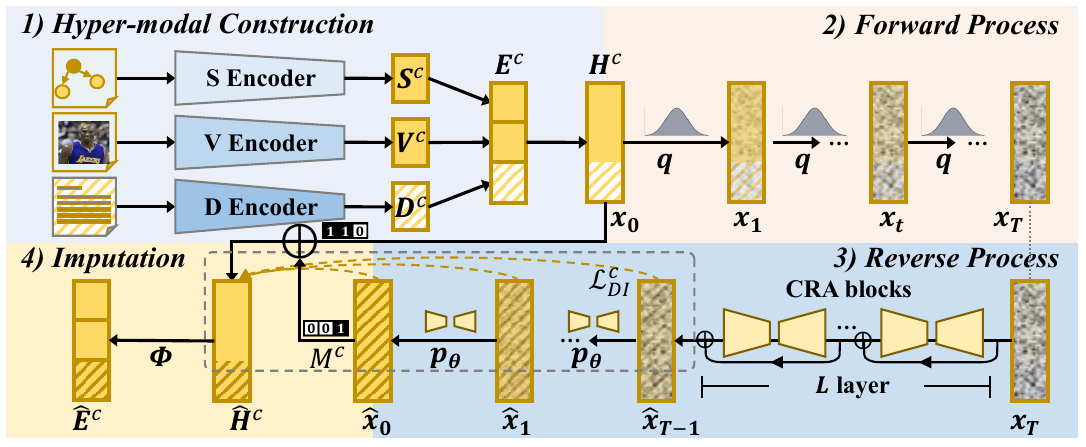}
                  \caption{The HidE model details, including hyper-modal construction, forward process, reverse process, and imputation. 
        }
         \label{fig:diff_method}
\vspace{-10pt}
         
\end{figure}

%% file: scripts/experiments.tex
\input{tables/dataset}

\input{tables/main_results}

\section{Experiments} \label{sec:experiments}

\subsection{Experimental Settings}
\textbf{Datasets.} We construct the FedMKGC benchmark as shown in Section \ref{sec:benchmark}.
We augment the dataset with the entity images \cite{wang2021rsme, liu2019mmkgdataset} and descriptions \cite{yao2019kgbert} from FB15K-237. 
The train/valid/test split ratio is 8:1:1. 
The dataset statistics are shown in Table \ref{tab: dataset}, where the client number, and the average number of relations, entities, and triples are reported.
We also report the average visual and text similarity $\overline{S}_m^c$ of the same entity between clients, with ViT embedding similarity and Jaccard similarity respectively.

\textbf{Evaluation metrics.}
We predict both head and tail entities and take the average metrics as the client MKG metrics \cite{bordes2013TransE}, then we report the weighted average metrics of all clients based on their triple number ratio \cite{chen2021FedE}. 
We report four metrics, i.e. Hits@K, K=1, 3, 10, and mean reciprocal rank (MRR). 
We employ the widely-used filtered evaluation setting \cite{bordes2013TransE} to filter out other ground-truth entities. 

\textbf{Implementation details.}
We implement our framework with PyTorch. 
We employ RotatE \cite{sun2018RotatE,zhang2024unleashing} as KGE decoder $f(\cdot)$.
We employ Adam \cite{kingma2014adamoptimization} to optimize the framework. 
At each round, we sample all the clients to form $\mathcal{C}_t$ \cite{chen2021FedE}. 
We train the local clients for 3 epochs in each round. 
We early stopped the training after 5 rounds without improvement on the validation set. 
The batch size is 1,024.
The negative sample set $N_i$ is 256 randomly sampled negative entities. 
The embedding sizes of entity and relation are 512 and 256 \cite{sun2018RotatE}. 
The diffusion noise $\beta_{low},\beta_{up}$ is $5e-4, 5e-2$ respectively, scaled with $s=1e-4$ \cite{wang2023diffrec}. 
Hyperparameters are tuned on the validation set. 
The parameter sensitivity analysis is shown in App. C.D. 
We ran our experiments on one A6000 GPU (48GB RAM).

\subsection{Baselines construction}
We experiment with baselines from three groups:
(1) \textbf{MKGC baselines} for $\Phi(\cdot)$: Average, Weighted \cite{wang2021rsme}, Concatenation \cite{wang2021rsme}, Split \cite{zhao2022mose,li2023imf}, Gated \cite{kiela2018gated}.
We first benchmark the MKGC baselines without missing modalities, shown in App. C.B, where weighted fusion approach shows superior performance. Thus, we conduct the following experiments with weighted as default $\Phi(\cdot)$. 
(2) \textbf{Federated learning (FL) baselines}: MMFedE \cite{chen2021FedE}, MMFedEC \cite{chen2022FedEC}, MMFedProx \cite{li2020FedProx}, MMFedLU \cite{zhu2023FedLU}, derived from our MMFedE backbone.
(3) \textbf{Incomplete multimodal learning (IMML) baselines}: AE \cite{baldi2012autoencoders}, CRA \cite{tran2017missingcra}, MMIN \cite{zhao2021MMIN}, adapted with our masked objective. 
The details of baseline construction are in App. C.A.

Moreover, we construct the lower and upper bound of FedMKGC.
The lower bound \textbf{S-Ind} denotes the average performance of independently trained structural models without FL and multimodal information, while  \textbf{MMInd} denotes that of independently trained multimodal models with partitioned images or descriptions.
The upper bound \textbf{MMCen} denotes the performance of a centralized trained global model, where all triples of client MKGs are available and multimodal information is not partitioned or missing.

\subsection{FedMKGC with missing modalities}

As shown in Table \ref{tab:mainresults}, we experiment with $r=50\%$ available modalities and construct the evaluation of FL and IMML baselines.
HidE outperforms IMML baselines and MMFeD3 outperforms FL baselines on 3 datasets, demonstrating the effectiveness of our method.

\textbf{Lower and upper bound: }
{\textit{Compared to the lower bound}}, MMFeD3-HidE outperforms S-Ind on three datasets significantly.
It demonstrates that MMFeD3-HidE could effectively utilize the federated learning and the partially available multimodal information to jointly learn better embeddings for MKGC.
{\textit{Compared to the upper bound}}, MMFeD3-HidE also outperforms MMCen-weighted on three datasets. 
The reason could be that MMFeD3-HidE is effective in imputing the uncertainly unavailable modalities to further improve FedMKGC performance. 

\textbf{FL methods comparison: }
MMFeD3 outperforms MMFedE, MMFedEC, MMFedProx, and MMFedLU on three datasets and with different incomplete multimodal learning methods, demonstrating the effectiveness of MMFeD3. 
With different multimodal learning methods, MMFeD3 co-distillates knowledge between predictions and embeddings of clients with reconstructed multimodal and the server with existing modalities. 
The reason could be that MMFeD3 is effective in guiding all the reconstruction networks, as well as guiding the global prediction accuracy.
Moreover, the dual distillation method could effectively deal with the heterogeneity between clients for stable global convergence to better performance.

\textbf{IMML methods comparison: }
HidE generally outperforms AE, CRA, and MMIN with different FL approaches on 3 datasets.
The reason could be that they lack an effective utilization of available modalities. 
Our HidE made full use of existing modalities by reserving them in the final $\hat{\mathbf{E}}^c$ through imputation and step-wise diffusion supervision. 
Moreover, the original reconstruction-based approaches in their paper \cite{tran2017missingcra,zhao2021MMIN} are optimized with ground-truth features, which could be constrained by our application scenario that can only exploit the self-supervision from the existing modalities inside the entity.
Without the supervised signal from the ground-truth feature, our proposed imputation diffusion model is beneficial for incomplete multimodal learning. 
Moreover, the diffusion model rebuilds the complete modalities step by step with step-wise supervision, ensuring the reconstruction stability and semantic consistency, leading to better performance.

\input{tables/ablation}

\input{tables/data_ablation}

\subsection{Ablation Study}
We conducted the model ablation study in Table \ref{tab: model_ablation} and data ablation study in Table \ref{tab:modality_ablation}. 

\subsubsection{Model Ablation Study}

\revision{
Ablating the diffusion imputation objective $\mathcal{L}_{DI}$ decreases the MRR by 4.3\%, showing the necessity of guiding diffusion reconstruction with masked supervision from observed modalities. 
Removing the feature distillation $\mathcal{L}_{FD}$ leads to performance drops of 1.1\% in MRR, verifying its contribution to maintaining semantic consistency during reconstruction and aggregation. 
With the server KGC loss $\mathcal{L}_{KGC}^{s,c}$ and logit distillation, the server and client performance are nearly identical as shown in Figure \ref{fig:fl_stable}. 
Ablating the server KGC loss brings a large downgrade of server performance, which also impacts the client performance, demonstrating the necessity to train the global embeddings simultaneously.
The removal of federated logit distillation objectives $\mathcal{L}_{LD} + \mathcal{L}_{KGC}^{s,c}$ results in a significant decay to 0.359 MRR, demonstrating the necessity of our dual-distillation mechanism for handling client heterogeneity and guiding global convergence. 
}

\revision{
\textbf{Structural Embedding Initialization. }
To investigate the sensitivity of HidE to structural quality, we evaluated variants initialized with random noise ("w/ random. init. $S^c$") and structural embeddings learned from only 50\% triples ("w/ $\frac{1}{2}\mathcal{T}^c$ init. $S^c$"). While the performance decreases to 0.371 and 0.375 MRR, respectively, due to the weakened structural priors, they still significantly outperform other IMML baselines, proving that our diffusion module remains robust and can effectively recover multimodal semantics even when the local structural information is sparse or noisy.
}

\subsubsection{Data Ablation Study}

\textbf{Complete v.s. Incomplete Multimodal: }\textit{In the full multimodal group} where $r=100\%$, MMFeD3 outperforms MMFedE by $1.8\%$ MRR with partitioned multimodal information, demonstrating the effectiveness of MMFeD3. 
\textit{In the uncertain incomplete multimodal group where} $r=50\%$, MMFeD3-HidE also outperforms MMFedE by $2.3\%$ MRR.
Moreover, after imputation, the performance of MMFeD3-HidE is similar to MMFeD3 without unavailable modalities, demonstrating the effectiveness of HidE for imputing the missing modalities.

\textbf{Modality ablation:} \textit{Comparing S+T+V with S+$r$D and S+$r$V}, the ablation of visual or textual modality brings little performance drop, demonstrating the stability of MMFedE and MMFeD3 in incorporating multiple modalities.
Moreover, the weighted fusion method may be too naive to promote reasoning ability from multimodal complementary, indicating future investigation directions of effective multimodal fusion methods for FedMKGC. 

\textbf{Multimodal partitioned v.s. non-partitioned:} 
The performance of MMFeD3-HidE with partitioned information is similar to that with non-partitioned one, which shows the robustness of our approach under heterogeneous non-IID multimodal information. 
It also shows that the reconstructed incomplete multimodal information of MMFeD3-HidE is comparable with the unpartitioned complete one.

\input{figs/visualization}

\subsection{Semantic Consistency Analysis}

\subsubsection{Visualization: }
We visualize entity embedding of one client in FB15K-237-Fed3 with t-SNE \cite{van2008tsne} as shown in Figure \ref{fig:vis}.
\textbf{Entity type: }
The same-type entity embeddings from MMFeD3-HidE are closer in a cluster than those from MMFedE, and the different-type ones are further, showing effectiveness of MMFeD3-HidE in capturing entity semantics to prevent entity ambiguity. 
The embeddings from MMFedE are more scattered, demonstrating the necessity of effective IMML and FL approach. 

\textbf{Missing type: }
Some of the MMFedE embeddings with missing modalities are away from the type cluster, while those of MMFeD3-HidE are all clustered close, showing that HidE could dynamically maintain the consistent semantics in entity representations regardless of their uncertain missing types. 

\revision{
\subsubsection{Quantitative semantic consistency: }
We evaluate Mean Squared Error (MSE) scores between diffusion-imputed embeddings and the observed embeddings on the test set of MMFeD3-HidE on FB15K-237-Fed3.
Specifically, MMFeD3-HidE achieves $3.59 \times 10^{-4}$ for the textual modality and $3.20 \times 10^{-4}$ for the visual modality, with an overall hyper-modal MSE of $13.7 \times 10^{-4}$. 
These error rates serve as strong evidence that the HidE module successfully models the complex multimodal data manifold, ensuring that the imputed features are not only visually coherent in t-SNE but also distributionally aligned with the ground-truth semantic space. 
}

\subsection{Federated Convergence Efficiency Analysis}
Figure \ref{fig:fl_stable} shows the validation performance of FL methods during training to demonstrate convergence stability and our FL efficiency.
Compared to MMFedE/MMFedEC/MMFedLU, MMFeD3 not only reaches better performance than the baselines, but also shows a faster convergence speed that reaches convergence in fewer iterations.
Compared with the same distillation-based method MMFedLU, the server and client of MMFeD3 can reach better performance through co-guidance with both logit and feature distillation. 
MMFedLU only exploits logit distillation and underperforms MMFeD3, demonstrating the necessity of feature distillation in further improving server-client semantic consistency. 
MMFedLU has better server performance than client performance, which is not as stable as MMFeD3.
On the contrary, the client performance of MMFeD3 is as high as the server, ensuring the local models have the same MKGC inference ability. 
\input{figs/fl_stable}

\revision{
We further investigate the computational overhead of HidE. 
First, regarding the trainable parameters, the size of AE, CRA, MMIN, and HidE are 1.90, 10.43, 20.86, and 10.45 (M), respectively. 
HidE is significantly more lightweight than the competitive IMML baseline MMIN and comparable to CRA. 
Regarding training time consumption, average seconds per iteration of AE, CRA, MMIN, and HidE with MMFeD3 are 152.6, 323.6, 550.6, and 1048.4, respectively.
While the multi-step diffusion mechanism inherently increases the computation time per iteration compared to single-step reconstruction methods, the diffusion process in HidE provides inference performance improvements, which is an acceptable trade-off between time and effectiveness. 
We have investigated the hyperparameters of diffusion steps in App. C.D., which shows that fewer steps would not significantly decrease performance but increase efficiency. 
Furthermore, regarding communication overhead, since the diffusion parameters are maintained locally, the HidE module imposes no additional transmission burden on the federated communication framework.
}

\input{figs/mmar_zhexian}

\subsection{Effect of Multimodal Available Rate} \label{supp:mm available rate}
\revision{
We study the effect of multimodal available rates $r$ of available entity images and descriptions. 
As Figure \ref{fig:mmar} shows, with a low available rate, the performance of MMFedE decreases in general, which shows the harm from uncertain missing modalities. 
With our MMFeD3 for stable federated optimization and with HidE to impute the uncertain missing modalities, the performance of MMFeD3-HidE is not only higher than MMFedE but also robust to multimodal available rates, so that MMFeD3-HidE performance is always significantly higher than MMFedE.
}

%% file: tables/dataset.tex
\begin{table}[!b]
\vspace{-15pt}
    
     \caption{Dataset statistics of Federated MKGC.}\label{tab: dataset} 
\vspace{-20pt}

    \begin{center}
    \resizebox{0.5\textwidth}{!}{
\begin{tabular}{lrrrrrr}
\toprule
Dataset  & $|\mathcal{C}|$ & $\overline{|\mathcal{R}^c|}$ & $\overline{|\mathcal{E}^c|}$ & $\overline{|\mathcal{T}^c|}$ &$\overline{S^c_{v}}$ &$\overline{S^c_{d}}$ \\

\midrule
FB15K-237-Fed3 & 3 & 79.0 & 12,595.3 & 103,373.0 & 0.331	&0.508  \\
FB15K-237-Fed5 & 5 & 47.4 & 11,260.4 & 62,023.2 & 0.334	&0.523 \\
FB15K-237-Fed10 & 10 & 23.7 & 8,340.1 & 31,011.6 & 0.333&	0.542 \\

\bottomrule
\end{tabular}
}
    \end{center}
    \end{table}

%% file: tables/main_results.tex
\begin{table*}[!t]
\centering
\caption{FedMKGC performance with uncertain missing entity images and descriptions. 
We experiment with $r=50\%$ available entity images and descriptions.
The S-Independent and MMCen-weighted are lower and upper bound, respectively. 
The default multimodal fusion method is weighted fusion.
}
\label{tab:mainresults}

\resizebox{\linewidth}{!}{
\setlength{\tabcolsep}{1.6mm}{\begin{tabular}{l|l|cccc|cccc|cccc}

\toprule
\multirow{2}{*}{IMML}&\multirow{2}{*}{FL} &  \multicolumn{4}{c|}{FB15K-237-Fed3} & \multicolumn{4}{c|}{FB15K-237-Fed5} & 
\multicolumn{4}{c}{FB15K-237-Fed10}\\
\cline{3-6}
\cline{7-10}
\cline{11-14}
& & Hits@1 & Hits@3 & Hits@10 & MRR & Hits@1 & Hits@3 & Hits@10 & MRR & Hits@1 & Hits@3 & Hits@10 & MRR\\

\midrule
\multirow{2}{*}{-}&S-Ind & 0.231 & 0.390 & 0.547 & 0.338 & 0.220 & 0.373 & 0.532 & 0.325 & 0.205 & 0.361 & 0.517 & 0.310\\
&MMCen-weighted  & {0.246} & {0.441} & {0.616} & {0.373} & 0.251 & {0.451} & {0.627} & {0.380}  & 0.240 & 0.447 & {0.629} & 0.373 \\
\midrule
\multirow{5}{*}{AE}&MMFedE & 0.133 & 0.284 & 0.458 & 0.241  &0.109 & 0.246 & 0.404 & 0.208 & 0.124 & 0.245 & 0.388 & 0.213 \\
&MMFedEC & 0.135 & 0.289 & 0.459 & 0.244&  0.111 & 0.247 & 0.409 & 0.210 & 0.124 & 0.238 & 0.382 & 0.211 \\
&MMFedProx   & 0.153 & 0.301 & 0.465 & 0.258&  \textbf{0.146} & 0.286 & 0.441 & 0.246 & 0.115 & 0.220 & 0.367 & 0.198 \\
&MMFedLU   & 0.160 & 0.317 & 0.489 & 0.270 & 0.140 & 0.276 & 0.435 & 0.238  &0.115 & 0.253 & 0.409 & 0.214 \\
&MMFeD3 & \textbf{0.173} & \textbf{0.354} & \textbf{0.521} & \textbf{0.293} & 0.144 & \textbf{0.299} & \textbf{0.451} & \textbf{0.250} &  \textbf{0.151} & \textbf{0.309} & \textbf{0.474} & \textbf{0.260} \\
\midrule
\multirow{5}{*}{CRA}&MMFedE & 0.167 & 0.346 & 0.522 & 0.287&  0.140 & 0.312 & 0.493 & 0.259 & 0.137 & 0.313 & 0.499 & 0.258 \\
&MMFedEC    & 0.173 & 0.352 & 0.529 & 0.293 & 0.140 & 0.309 & 0.488 & 0.257 & 0.136 & 0.309 & 0.494 & 0.256 \\
&MMFedProx  & 0.214 & 0.408 & 0.580 & 0.340 & 0.198 & 0.390 & 0.560 & 0.324 & 0.155 & 0.339 & 0.520 & 0.279 \\
&MMFedLU  & 0.072 & 0.347 & 0.554 & 0.242 & 0.140 & 0.304 & 0.472 & 0.253 & 0.087 & 0.316 & 0.504 & 0.232 \\
&MMFeD3  & \textbf{0.237} & \textbf{0.419} & \textbf{0.590} & \textbf{0.357} & \textbf{0.268} & \textbf{0.428} & \textbf{0.583} & \textbf{0.376} & \textbf{0.238} & \textbf{0.385} & \textbf{0.540} & \textbf{0.341} \\
\midrule
\multirow{5}{*}{MMIN}&MMFedE    & 0.157 & 0.327 & 0.505 & 0.274 & 0.135 & 0.304 & 0.481 & 0.251 & 0.121 & 0.290 & 0.472 & 0.239 \\
&MMFedEC   & 0.160 & 0.329 & 0.506 & 0.276 & 0.140 & 0.297 & 0.469 & 0.250 & 0.125 & 0.291 & 0.472 & 0.242 \\
&MMFedProx & 0.208 & 0.397 & 0.566 & 0.332 & 0.199 & 0.389 & 0.557 & 0.323 & 0.158 & 0.345 & 0.521 & 0.283 \\
&MMFedLU & 0.139 & 0.298 & 0.465 & 0.249 & 0.136 & 0.298 & 0.470 & 0.248 & 0.131 & 0.280 & 0.445 & 0.236 \\
&MMFeD3 & \textbf{0.233} & \textbf{0.413} & \textbf{0.581} & \textbf{0.353} & \textbf{0.269} & \textbf{0.431} & \textbf{0.589} & \textbf{0.378} & \textbf{0.237} & \textbf{0.392} & \textbf{0.548} & \textbf{0.343} \\
\midrule
\multirow{5}{*}{HidE}&MMFedE    & 0.259 & 0.447 & 0.612 & 0.381 & 0.252 & 0.449 & 0.620 & 0.379 & 0.237 & 0.422 & 0.609 & 0.369 \\
&MMFedEC   & 0.260 & 0.446 & 0.613 & 0.382 & 0.251 & \textbf{0.450} & 0.621 & 0.379 & 0.237 & 0.423 & 0.609 & 0.369 \\
&MMFedProx & 0.258 & 0.446 & 0.613 & 0.380 & 0.255 & \textbf{0.450} & 0.618 & 0.381 & 0.239 & 0.424 & 0.609 & 0.370 \\
&MMFedLU & 0.255 & 0.431 & 0.601 & 0.372 & 0.246 & 0.441 & 0.616 & 0.372 & 0.239 & 0.436 & 0.609 & 0.366 \\

&MMFeD3 & \textbf{0.269} & \textbf{0.449} & \textbf{0.615} & \textbf{0.387} & \textbf{0.260} & \textbf{0.450} & \textbf{0.622} & \textbf{0.382} & \textbf{0.246} & \textbf{0.439} & \textbf{0.614} & \textbf{0.372}\\

\bottomrule

\end{tabular}
}
}

\end{table*}

%% file: tables/ablation.tex
\begin{table}[!t]
\centering
\caption{Model Ablation study of MMFeD3-HidE.
}
\label{tab: model_ablation}
\resizebox{0.95\linewidth}{!}{
\setlength{\tabcolsep}{2mm}{
\begin{tabular}{lcccc}
\toprule
\multirow{2}{*}{} & \multicolumn{4}{c}{FB15K-237-Fed3}  \\
\cline{2-5}
&  H@1&H@3 & H@10 & MRR \\
\midrule
MMFeD3-HidE & 0.269 & 0.449 & 0.615 & 0.387 \\
~~ w/o $\mathcal{L}_{DI}$ &
0.221&	0.407	&0.578&	0.344\\
~~ w/o $\mathcal{L}_{FD}$ & 0.256 &	0.437 &	0.608	& 0.376 \\
\midrule
~~ w/o $\mathcal{L}_{KGC}^{s,c}$ (client) & 0.267 & 0.442 & 0.611 & 0.384 \\
~~ w/o $\mathcal{L}_{KGC}^{s,c}$ (server) & 0.249 & 0.430 & 0.601 & 0.369 \\
~~ w/o $\mathcal{L}_{LD}+\mathcal{L}_{KGC}^{s,c}$ & 0.238 &	0.418 &	0.597 &	0.359 \\
\midrule
~~ w/ random. init. $\mathbf{S}^c$ & 0.258 &	0.425 &	0.596 &	0.371 \\
~~ w/ $\frac{1}{2} \mathcal{T}^c$ init. $\mathbf{S}^c$ & 0.257 & 0.435 & 0.605 & 0.375 \\

\bottomrule
\end{tabular}}}
\vspace{-10pt}
\end{table}

%% file: tables/data_ablation.tex
\begin{table*}[!t]
\centering

\caption{
Modality ablation results on FB15K-237-Fed3. S, D, and V represent structural, textual, and visual modalities, respectively. \textit{MM Partitioned} denotes whether the entity texts and images between clients are different. $r$ represents the visual/textual available rate.
}
\label{tab:modality_ablation}
\resizebox{\linewidth}{!}{
\setlength{\tabcolsep}{1.4mm}{\begin{tabular}{l|l|c|cccc|cccc|cccc}
\toprule
\multirow{2}{*}{$r$}&\multirow{2}{*}{Model} & \multirow{2}{*}{\makecell[c]{MM\\Partitioned}}  & \multicolumn{4}{c|}{S + $r$D} & \multicolumn{4}{c|}{S + $r$V} &
\multicolumn{4}{c}{S + $r$D + $r$V} \\
\cline{4-7}
\cline{8-11}
\cline{12-15}
&&& Hits@1  & Hits@3  & Hits@10  & MRR  &  Hits@1  & Hits@3  & Hits@10 &  MRR  &  Hits@1  & Hits@3  & Hits@10  & MRR  \\
\midrule
\multirow{4}{*}{100\%}&\multirow{2}{*}{MMFedE} & \xmark & 0.241 & 0.441 & 0.614 & 0.369 &   0.246 & 0.441 & 0.611 & 0.372 &   0.243          & 0.443          & 0.614 & 0.371          \\
&& \cmark &0.239 & 0.442 & 0.615 & 0.369 &   0.242 & 0.437 & 0.608 & 0.368 &   0.237          & 0.437          & 0.612 & 0.367          \\
\cline{2-15}
&\multirow{2}{*}{MMFeD3} & \xmark & 0.264 & \textbf{0.448} & \textbf{0.618} & \textbf{0.385} &   0.264 & \textbf{0.449} & \textbf{0.618} & \textbf{0.386} &   \textbf{0.268}          & \textbf{0.450}          & \textbf{0.618} & \textbf{0.387}          \\
&& \cmark  & \textbf{0.265} & 0.446 & 0.616 & \textbf{0.385} &   \textbf{0.265} & 0.446 & 0.616 & 0.385 &   0.265          & 0.446          & 0.616 & 0.385          \\
\midrule
\multirow{4}{*}{50\%}&\multirow{2}{*}{MMFedE} & \xmark  & 0.240 & 0.433 & 0.600 & 0.365 &   0.238 & 0.432 & 0.601 & 0.363 &   0.243          & 0.437          & 0.604 & 0.368          \\
&&\cmark  & 0.239 & 0.432 & 0.600 & 0.364 &   0.236 & 0.432 & 0.600 & 0.362 &   0.238          & 0.432          & 0.601 & 0.364          \\
\cline{2-15}
&\multirow{2}{*}{MMFeD3-HidE} & \xmark  & \textbf{0.261} & \textbf{0.441} & 0.614 & \textbf{0.381} &   \textbf{0.260} & 0.442 & \textbf{0.615} & \textbf{0.381} &   0.262          & 0.443          & 0.611 & 0.381          \\
&& \cmark  & \textbf{0.261} & \textbf{0.441} & \textbf{0.615} & \textbf{0.381} &   0.259 & \textbf{0.445} & \textbf{0.615} & 0.380 &   \textbf{0.269} & \textbf{0.449} & \textbf{0.615}& \textbf{0.387}\\
\bottomrule
\end{tabular}
}

}
\end{table*}

%% file: figs/visualization.tex
\begin{figure}[t]
     \centering
         \centering
         \includegraphics[width=0.48\textwidth]{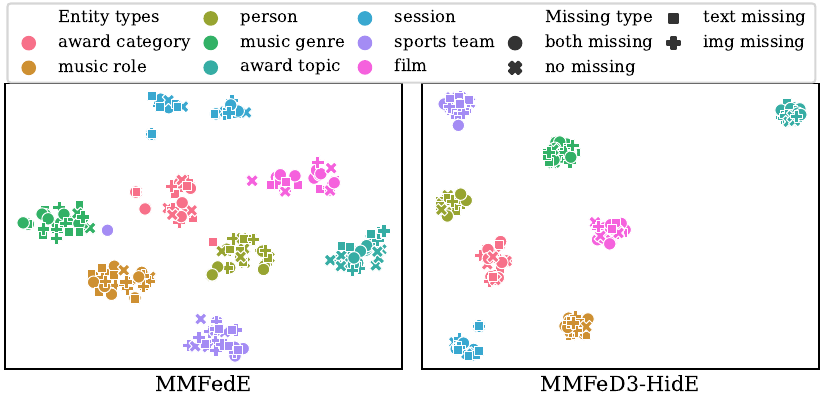}
         \caption{The t-SNE visualization of entity embeddings of MMFedE (w/o imputation) and MMFeD3-HidE. The entities have different types and different missing modalities.}
    \vspace*{\fill}
         \vspace{-16pt}
    
\label{fig:vis}
\end{figure}

%% file: figs/fl_stable.tex
\begin{figure}[t]
     \centering
         \centering
         \includegraphics[width=0.48\textwidth]{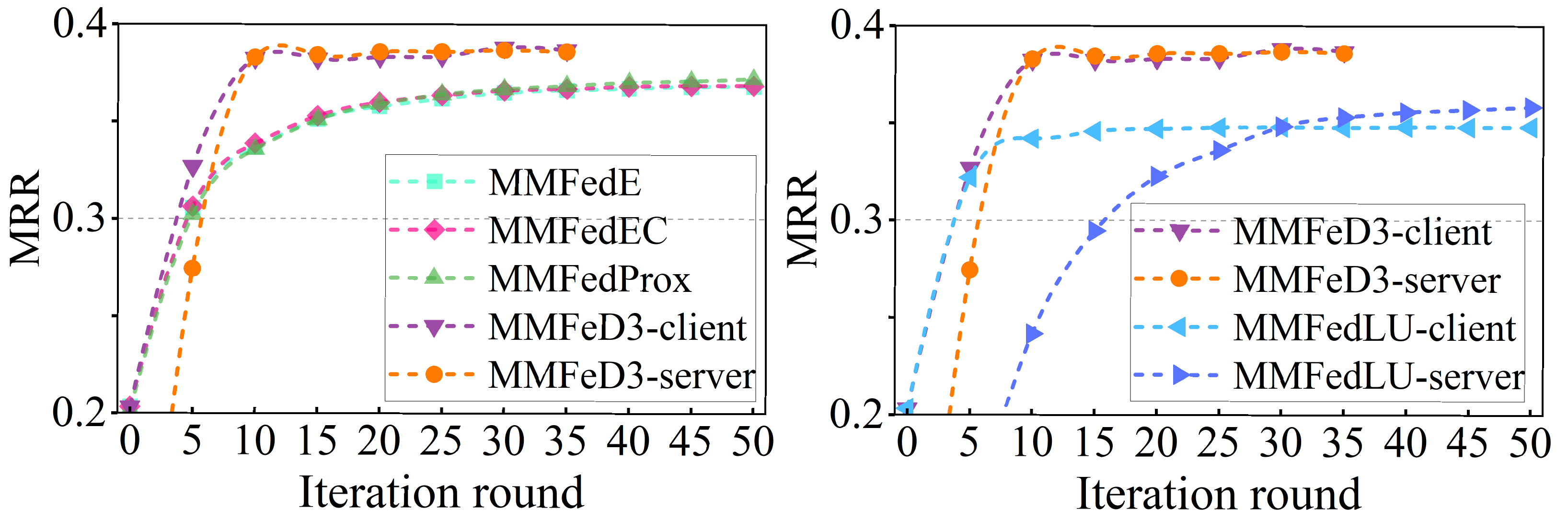}
        
         \caption{FL baseline performance on valid set during iteration.}
         \vspace{-10pt}
    \vspace*{\fill}
\label{fig:fl_stable}
\end{figure}

%% file: figs/mmar_zhexian.tex
\begin{figure}[t]
     \centering
         \centering
         \includegraphics[width=0.45\textwidth]{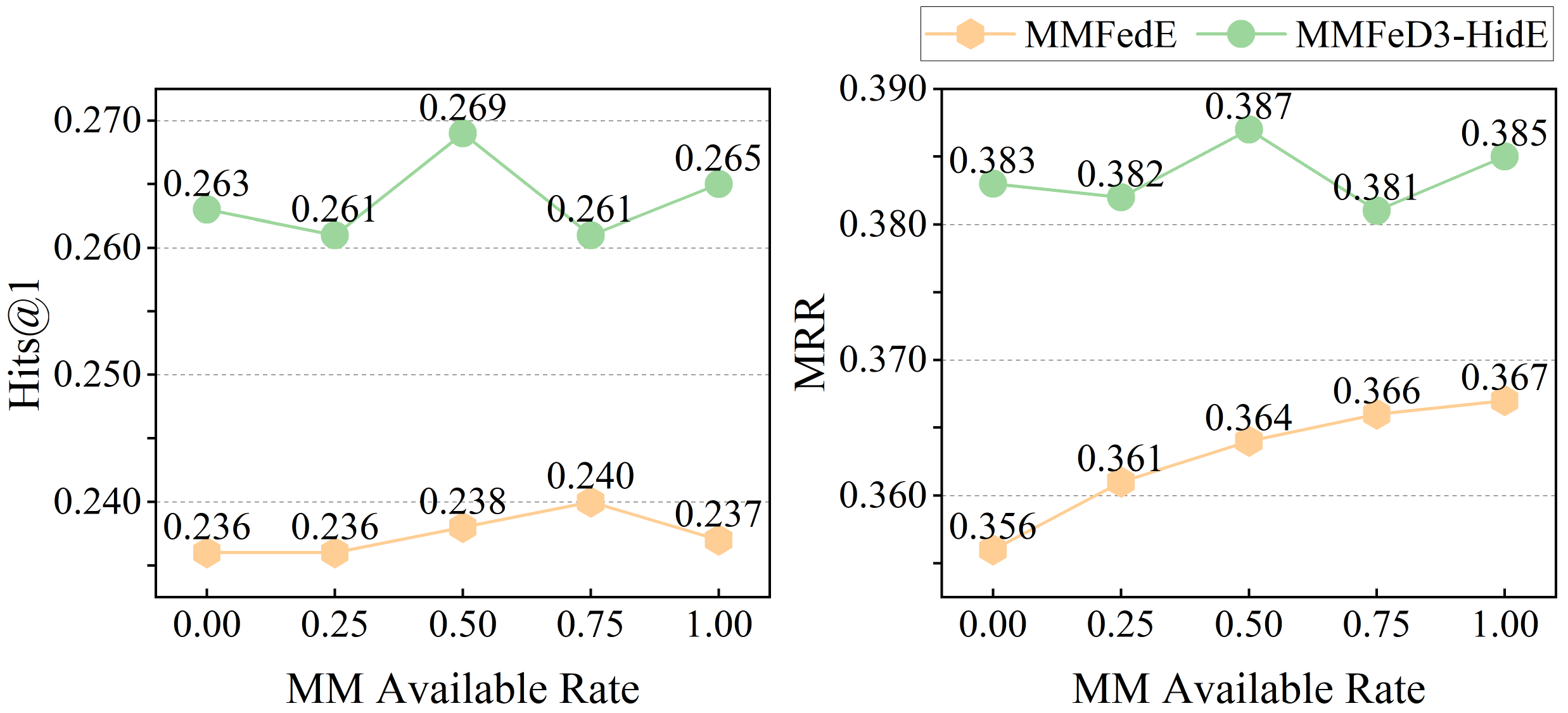}
         \caption{Performance comparison between MMFedE and MMFeD3-HidE with various multimodal available rates $r$.}
    \vspace*{\fill}
    \vspace{-15pt}
\label{fig:mmar}
\end{figure}

%% file: scripts/conclusion.tex
\section{Conclusion} \label{sec:conclusion}
In this paper, we propose the Federated Multimodal Knowledge Graph Completion (FedMKGC) task 
to collaboratively learn to complete the missing links on client MKGs without data transmission.
We explore FedMKGC challenges including multimodal uncertain unavailability and multimodal client heterogeneity.
We propose a MMFeD3-HidE framework to impute incomplete multimodal entity embeddings for multimodal semantic integrity and client-server dual distillation for robust federated optimization.
Moreover, we propose a FedMKGC benchmark with datasets, a general backbone, and three groups of baselines.
Extensive experiments validate the effectiveness, semantic consistency, and convergence efficiency of MMFeD3-HidE.
Future works include FedMKGC with provable privacy-preserving, effective multimodal fusion, and adaptive cross-institute communication.

%% file: photos/authors.tex
\begin{IEEEbiography}
[{\includegraphics[width=1in,height=1.25in,clip,keepaspectratio]{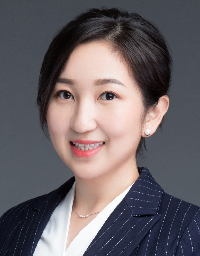}}]{Ying Zhang}
received the Ph.D. degree in computer science from Nankai University, China, in 2013. From 2011 to 2013, she studied at Purdue University, West Lafayette, IN, USA. She is currently a Professor with the College of Computer Science, Nankai University. Her main research interests include multimodal knowledge graphs, multimodal data analysis, and information retrieval.
\end{IEEEbiography}
\vspace{-30pt}
\begin{IEEEbiography}[{\includegraphics[width=1in,height=1.25in,clip,keepaspectratio]{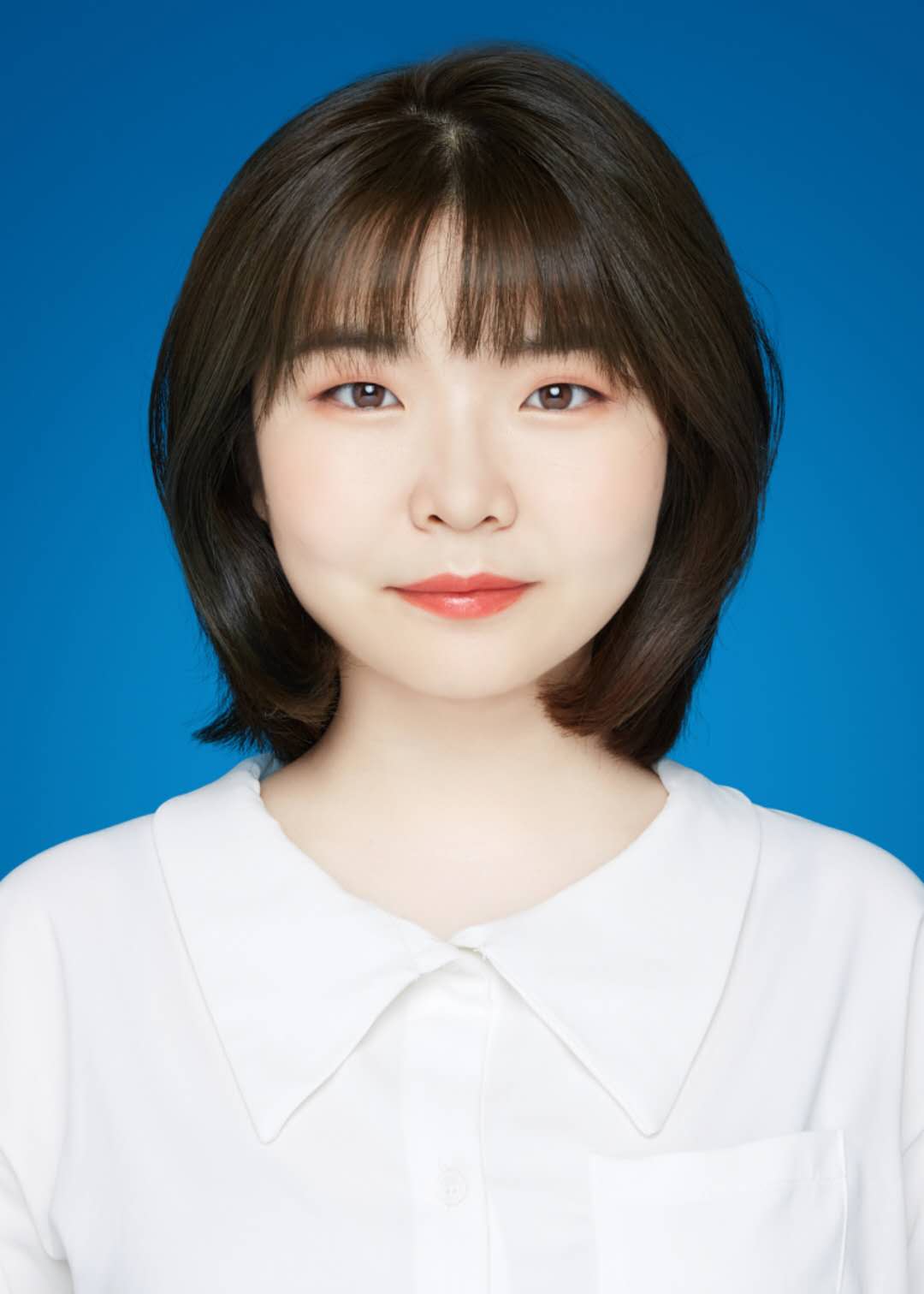}}]{Yu Zhao}
received the B.S. degree in software engineering from Nankai University, China, in 2021. She is currently pursuing the Ph.D. degree with the College of Computer Science, Nankai University, China. 
From 2025 to 2026, she studies at Nanyang Technological University (NTU), Singapore.
Her current research interests include multimodal knowledge graphs, knowledge representation and reasoning, and large language models.
\end{IEEEbiography}
\vspace{-30pt}
\begin{IEEEbiography}[{\includegraphics[width=1in,height=1.25in,clip,keepaspectratio]{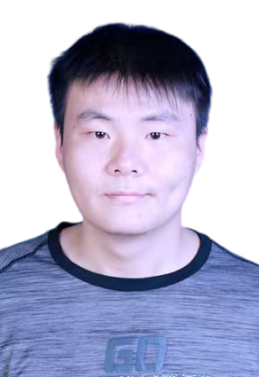}}]{Xuhui Sui}
received the B.S. degree in computer science from Nanchang University, China, in 2020. He is currently pursuing the Ph.D. degree with the College of Computer Science, Nankai University, China. His current research interests include multimodal knowledge graphs, entity linking, and data mining.
\end{IEEEbiography}
\vspace{-30pt}
\begin{IEEEbiography}[{\includegraphics[width=1in,height=1.25in,clip,keepaspectratio]{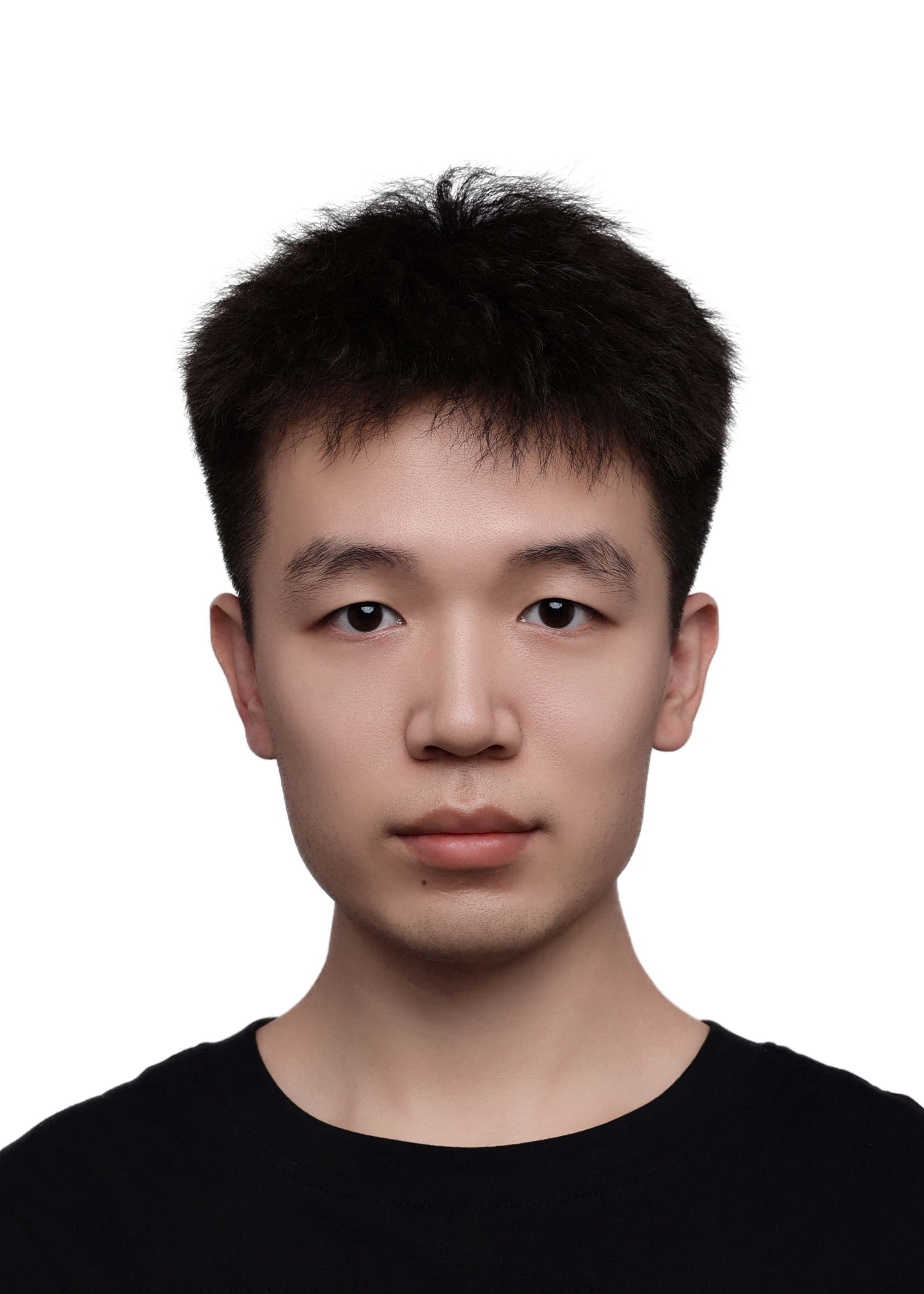}}]{Baohang Zhou}
received the Ph.D. degree in computer science from Nankai University, China, in 2025. From 2023 to 2024, he studied at Chinese University of Hong Kong, China. He is currently a Lecturer with the School of Software, Tiangong University. His main research interests include multimodal knowledge graphs, data mining, and large language model.
\end{IEEEbiography}
\vspace{-30pt}
\begin{IEEEbiography}[{\includegraphics[width=1in,height=1.25in,clip,keepaspectratio]{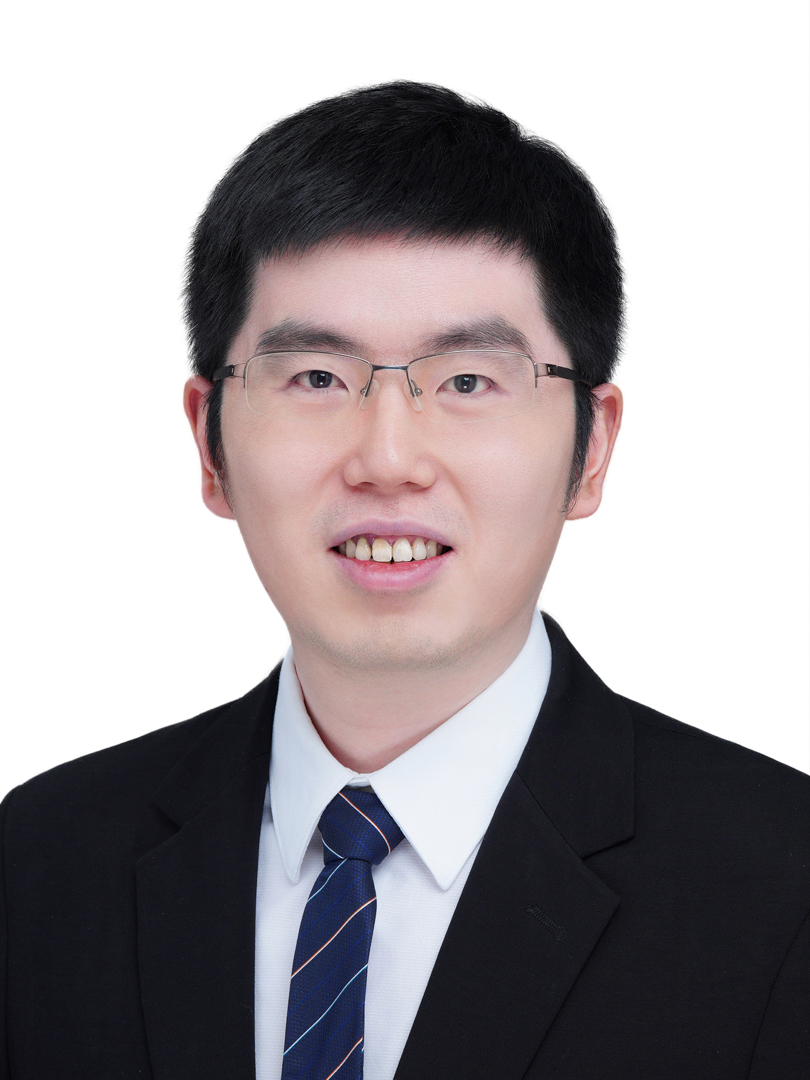}}]{Xiangrui Cai}
received the Ph.D. degree in computer science from Nankai University, China. He is currently working as an associate professor at the College of Computer Science, Nankai University. His main research interests include learning with missing data, natural language processing, and AI safety.
\end{IEEEbiography}
\vspace{-30pt}
\begin{IEEEbiography}[{\includegraphics[width=1in,height=1.25in,clip,keepaspectratio]{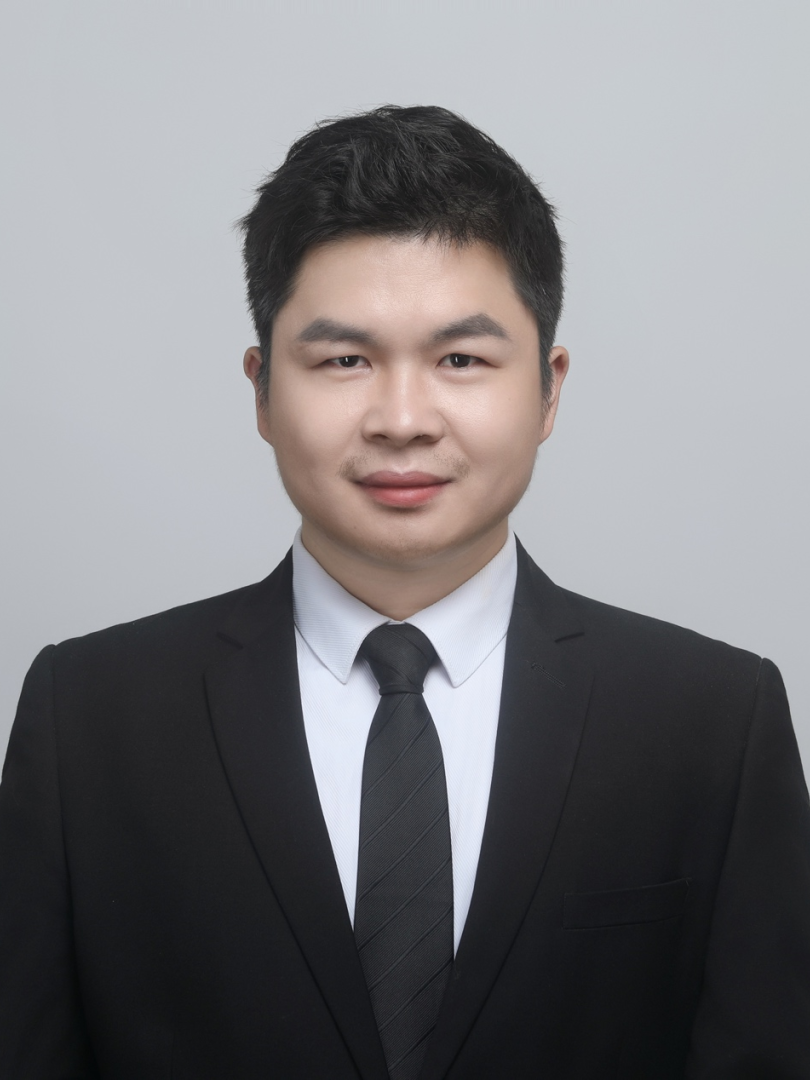}}]{Li Shen}
received the bachelor’s and PhD degrees from the School of Mathematics, South China University of Technology. He is currently an associate professor with Sun Yat-sen University. Previously, he was a research scientist with JD Explore Academy, Beijing, and a senior researcher with Tencent AI Lab, Shenzhen. His research interests include theory and algorithms for nonsmooth convex and nonconvex optimization, and their applications in trustworthy artificial intelligence, deep learning, and reinforcement learning. He has published more than 100 papers in peer-reviewed top-tier journal papers (Journal of Machine Learning Research, IEEE Transactions on Pattern Analysis and Machine Intelligence, International Journal of Computer Vision, IEEE Transactions on Signal Processing, IEEE Transactions on Image Processing, IEEE Transactions on Knowledge and Data Engineering, etc.) and conference papers (ICML, NeurIPS, ICLR, CVPR, ICCV, etc.). He has also served as the senior program committee for AAAI and area chairs for ICML, ICLR, ACML, and ICPR.\end{IEEEbiography}
\vspace{-30pt}
\begin{IEEEbiography}[{\includegraphics[width=1in,height=1.25in,clip,keepaspectratio]{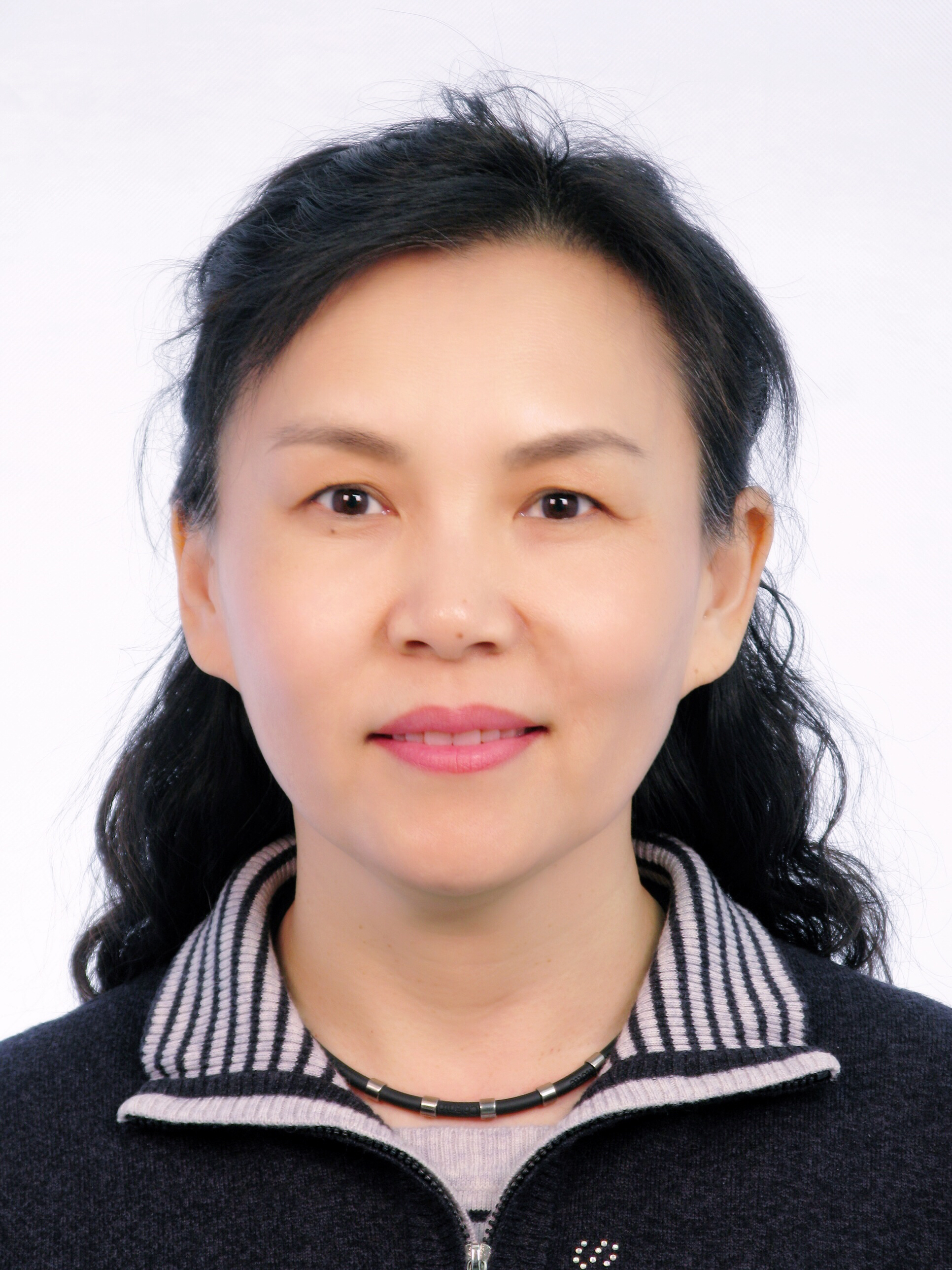}}]{Xiaojie Yuan}
received the Ph.D. degree in computer science from Nankai University, China, in 2000. She is currently a Professor with the College of Computer Science, Nankai University. She leads a research group working on topics of database, data mining, and information retrieval.
\end{IEEEbiography}
\vspace{-30pt}
\begin{IEEEbiography}[{\includegraphics[width=1in,height=1.25in,clip,keepaspectratio]{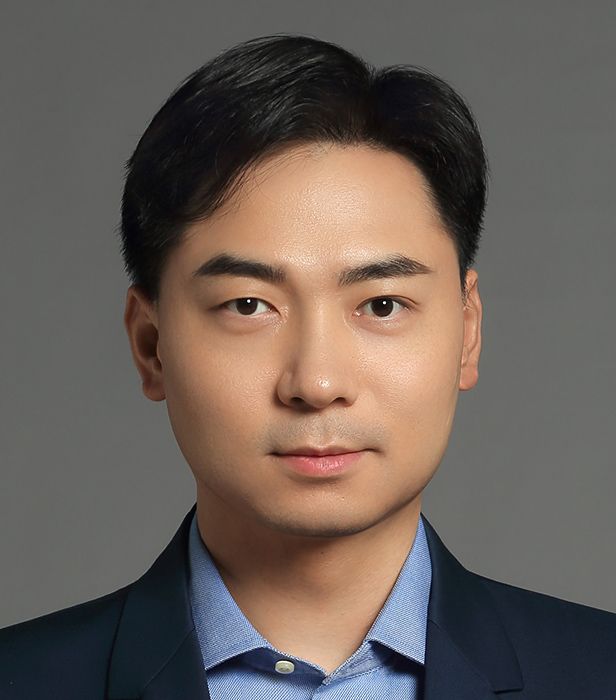}}]{Dacheng Tao}
(Fellow, IEEE) is currently a Distinguished University Professor with the College of Computing and Data Science, Nanyang Technological University. He mainly applies statistics and mathematics to artificial intelligence and data science, and his research is detailed in one monograph and more than 200 publications in prestigious journals and proceedings at leading conferences, with best paper awards, best student paper awards, and test-of-time awards. His publications have been cited more than 112 K times, and he has an H-index more than 160 in Google Scholar. He received the 2015 and 2020 Australian Eureka Prize, the 2018 IEEE ICDM Research Contributions Award, the 2019 Diploma of the Polish Neural Network Society, and the 2021 IEEE Computer Society McCluskey Technical Achievement Award. He is a fellow of the Australian Academy of Science, AAAS, and ACM.
\end{IEEEbiography}
\vspace{-30pt}

%% file: scripts/appendix.tex
\section{Notations} \label{supp:notations}
The notations used in this paper are shown in Table \ref{tab: notations}.
\input{tables/notations}

\section{Method Details}

\subsection{MMFeD3-HidE Pipeline} \label{supp:alg MMFeD3}
\input{algs/alg_mmfedcd}
The pipeline of MMFeD3-HidE is shown in Algorithm \ref{alg: pipeline hide-mmfed3}.

\subsection{Detailed Diffusion Imputation Optimization} \label{supp:ELBO detailed}

Diffusion models are typically optimized by maximizing the evidence lower bound (ELBO) of the likelihood of the original state $\mathbf{x}_0$ as Equation \eqref{eq: ELBO loss}. 
\begin{equation}\label{eq: ELBO loss}
\small
\begin{aligned}
    \log p(\mathbf{x}_0)&= \log \int p(\mathbf{x}_{0:T}) \mathrm{d}\mathbf{x}_{1:T} \\
    =& \log \mathbb{E}_{q(\mathbf{x}_{1:T}|\mathbf{x}_0)} \left[ \frac{p(\mathbf{x}_{0:T})}{q(\mathbf{x}_{1:T}|\mathbf{x}_0)} \right] \\
    \ge&  {\underbrace{\mathbb{E}_{q(\mathbf{x}_{1}|\mathbf{x}_0)}\left[\log p_{\theta}(\mathbf{x}_0|\mathbf{x}_1)\right]}_{\text{reconstruction term}}}- {\underbrace{ D_{\text{KL}}({q(\mathbf{x}_T|\mathbf{x}_{0})}\parallel{p(\mathbf{x}_T)})}_{\text{prior matching term}}} \\
      -& {\sum_{t=2}^{T}\underbrace{\mathbb{E}_{q(\mathbf{x}_{t}|\mathbf{x}_0)}\left[ D_{\text{KL}}({q(\mathbf{x}_{t-1}|\mathbf{x}_{t},\mathbf{x}_0)}\parallel{p_{\theta}(\mathbf{x}_{t-1}|\mathbf{x}_{t})})\right]}_{\text{denoising matching term}}}\\
\end{aligned}
\end{equation}

Based on DDPM, ELBO can be simplified as $\mathbb{E}_{t,\mathbf{x}_0,\boldsymbol{\epsilon}}\left[\parallel\boldsymbol{\epsilon} - \boldsymbol{\epsilon}_{\theta}(\mathbf{x}_t,t)\parallel_2^2\right]$, where $1\leq t\leq T$. 
\revision{DDPM's reconstruction network outputs the additional noise $\boldsymbol{\epsilon}_{\theta}$ and optimizes the noise to approximate the distribution step by step.
In our paper, we aim to predict the complete hyper-modal distributions conditioned on the input observed modalities; thus, we predict the complete hyper-modal $\mathbf{x}_0$ at each step instead of $\boldsymbol{\epsilon}_{\theta}$. 
Thus, in our HidE, the reconstruction network $\theta$ is optimized to regulate its output $\hat{\mathbf{x}}_\theta(\mathbf{x}_t, t)$ closer to $\mathbf{x}_0$.
This way, the output of the diffusion model can be directly constrained by the observed modalities as strong supervision signals and optimized to remain semantically consistent with the observed modalities. }

However, there are both missing and observed data points in the diffusion processes; we need to design the objective with reconstruction supervision signals that are not affected by the missing modal points. 
Since the diffusion model preserves the feature size and location in both forward and backward processes, it offers a convenient solution for diffusion imputation optimization.
We split the data at each step $\mathbf{x}_t$ to observed data and missing data $\mathbf{x}_t=\{\mathbf{x}_t^{obs},\mathbf{x}_t^{mis}\}$, that $\mathbf{x}_t^{obs}=M^c \odot \mathbf{x}_t$ and $\mathbf{x}_t^{mis}=(1-M^c) \odot \mathbf{x}_t,1\leq t\leq T$.
We propose to constrain the diffusion of available modalities $\mathbf{x}_t^{obs}$ to be similar to their ground-truth features.
\revision{This way, the output of the diffusion model is sampled from the same underlying conditional distribution $p(\mathbf{x}^{complete}|\mathbf{x}^{obs}_0)$, which is the complete hypermodal distribution conditioned on the observed modalities. 
This way, the data points of the missing modalities $\mathbf{x}_t^{mis}$ are also sampled from the conditional distribution $p(\mathbf{x}^{complete}|\mathbf{x}^{obs}_0)$, 
ensuring that the reconstructed hyper-modal embeddings maintain distributional consistency with the original MKGs with missing modalities.}
We elaborate on simplifying ELBO to masked diffusion imputation by analyzing the optimization of each term as follows:

\subsubsection{{Prior matching term optimization}}
The prior matching term measures the KL divergence between the posterior probability $q(\mathbf{x}_T|\mathbf{x}_{0})$ of $\mathbf{x}_T$ at step $T$ of $\mathbf{x}_T$ and the prior probability $p(\mathbf{x}_T)$. 
The prior matching term is omitted as constants since the forward process has no trainable parameters.

\subsubsection{{Denoising matching term optimization}}
The denoising matching term constrains the reconstruction $p_{\theta}(\mathbf{x}_{t-1}|\mathbf{x}_{t})$ at each step from $2$ to $T$, exploiting KL divergence to force the generated $\mathbf{x}_{t-1}$ to approximate forward process posteriors $q(\mathbf{x}_{t-1}|\mathbf{x}_{t},\mathbf{x}_0)$.
The approximations are tractable when conditioned on the available modalities in hyper-modal features $\mathbf{x}_0$.
According to Bayes rules, the forward process posteriors $q(\mathbf{x}_{t-1}|\mathbf{x}_{t},\mathbf{x}_0)$ can be rewritten as Equation \eqref{eq: q_cond}, where $\tilde{\boldsymbol{\mu}}_t(\mathbf{x}_t$ and $\tilde{\beta}_t$ are its mean and variance. 
\begin{equation} \label{eq: q_cond}
\small
    q(\mathbf{x}_{t-1}|\mathbf{x}_{t},\mathbf{x}_0) = \mathcal{N}(\mathbf{x}_{t-1}; \tilde{\boldsymbol{\mu}}_t(\mathbf{x}_t, \mathbf{x}_0), \tilde{\beta}_t\mathbf{I}) 
\end{equation}
\begin{equation} \label{eq: tilde_mu_beta_t}
\small
\left\{
\begin{aligned}
&\tilde{\boldsymbol{\mu}}_t(\mathbf{x}_t,\mathbf{x}_0) =\dfrac{\sqrt{\alpha_t}(1-\bar{\alpha}_{t-1})}{1-\bar{\alpha}_t}\mathbf{x}_t+\dfrac{\sqrt{\bar{\alpha}_{t-1}}(1-\alpha_t)}{1-\bar{\alpha}_t}\mathbf{x}_0 \\
&\tilde{\beta}_t =\dfrac{1-\bar{\alpha}_{t-1}}{1-\bar{\alpha}_t}\beta_t
\end{aligned}
\right.
\end{equation}

For reverse process, we first set the learning of $\mathbf{\Sigma}_\theta(\mathbf{x}_t,t)$ in reverse process to untrained time-dependent constants as $\mathbf{\Sigma}_\theta(\mathbf{x}_t,t) = \sigma_t^2 \mathbf{I}$ and  $\sigma_t^2=\tilde{\beta}_t$.
Thus, we define the loss of denoising matching term of ELBO as $\mathcal{L}_t$ at step $t$ as Equation \eqref{eq: L_t_denoise_match_term}, to push the $\boldsymbol{\mu}_\theta(\mathbf{x}_t, t)$ closer to $\tilde{\boldsymbol{\mu}}_t(\mathbf{x}_t, \mathbf{x}_0)$. 
By minimizing $\mathcal{L}_t$, the denoising matching term of ELBO is maximized.
\begin{equation} \label{eq: L_t_denoise_match_term}
\small
\begin{aligned}
\mathcal{L}_t &\triangleq \mathbb{E}_{q(\mathbf{x}_{t}|\mathbf{x}_0)}\left[ D_{\text{KL}}({q(\mathbf{x}_{t-1}|\mathbf{x}_{t},\mathbf{x}_0)}\parallel{p_{\theta}(\mathbf{x}_{t-1}|\mathbf{x}_{t})})\right] \\
&= \mathbb{E}_{q(\mathbf{x}_{t}|\mathbf{x}_0)}\left[ 
\dfrac{1}{2\sigma_t^2}\left[\parallel \boldsymbol{\mu}_\theta(\mathbf{x}_t, t) - \tilde{\boldsymbol{\mu}}_t(\mathbf{x}_t, \mathbf{x}_0) \parallel_2^2 \right]
\right]
\end{aligned}
\end{equation}

Secondly, we denote $\boldsymbol{\mu}_\theta(\mathbf{x}_t, t)$ in the same form in Equation \eqref{eq: tilde_mu_beta_t} for further simplicity as Equation \eqref{eq: mu_theta}, 
\begin{equation} \label{eq: mu_theta}
\small
\boldsymbol{\mu}_\theta(\mathbf{x}_t, t) = \dfrac{\sqrt{\alpha_t}(1-\bar{\alpha}_{t-1})}{1-\bar{\alpha}_t}\mathbf{x}_t+\dfrac{\sqrt{\bar{\alpha}_{t-1}}(1-\alpha_t)}{1-\bar{\alpha}_t} \hat{\mathbf{x}}_\theta(\mathbf{x}_t, t),
\end{equation}
where $\hat{\mathbf{x}}_\theta(\mathbf{x}_t, t)$ is the estimated hyper-modal features $\mathbf{x}_0$ based on $\mathbf{x}_t$ and step $t$, which is the output of the reconstruction network.
Thus, the $\mathcal{L}_t$ can be denoted as Equation \eqref{eq: L_t_x_theta}.
\begin{equation} \label{eq: L_t_x_theta}
\small
\begin{aligned}
\mathcal{L}_t &= \mathbb{E}_{q(\mathbf{x}_{t}|\mathbf{x}_0)}\left[ \dfrac{1}{2\sigma_t^2}\left[\parallel \boldsymbol{\mu}_\theta(\mathbf{x}_t, t) - \tilde{\boldsymbol{\mu}}_t(\mathbf{x}_t, \mathbf{x}_0) \parallel_2^2 \right] \right] \\
=&\mathbb{E}_{q(\mathbf{x}_{t}|\mathbf{x}_0)}\left[ \dfrac{1}{2\sigma_t^2} \dfrac{\sqrt{\bar{\alpha}_{t-1}}(1-\alpha_t)}{1-\bar{\alpha}_t} \parallel \hat{\mathbf{x}}_\theta(\mathbf{x}_t, t) - \mathbf{x}_0 \parallel_2^2  \right] \\
\Leftrightarrow & \mathbb{E}_{q(\mathbf{x}_{t}|\mathbf{x}_0)} \left[ \parallel \hat{\mathbf{x}}_\theta(\mathbf{x}_t, t) - \mathbf{x}_0 \parallel_2^2  \right] \\
\end{aligned}
\end{equation}

Since we only exploit observed modalities for optimization, we re-define $\mathcal{L}_t$ as Equation \eqref{eq: L_t_imputation}, 
\begin{equation} \label{eq: L_t_imputation}
\small
\begin{aligned}
\mathcal{L}_t& =\mathbb{E}_{q(\mathbf{x}_{t}|\mathbf{x}_0)} \left[ \parallel \hat{\mathbf{x}}^{obs}_\theta(\mathbf{x}_t, t) - \mathbf{x}^{obs}_0 \parallel_2^2  \right]\\
&= \mathbb{E}_{q(\mathbf{x}_{t}|\mathbf{x}_0)} \left[ \parallel M^c\odot\hat{\mathbf{x}}_\theta(\mathbf{x}_t, t) - M^c\odot\mathbf{x}_0 \parallel_2^2  \right],
\end{aligned}
\end{equation}
where only the available modalities and their reconstructed features are pulled together.

\subsubsection{{Reconstruction term optimization}}
The reconstruction term constrains the transition from the hidden state in the first step $\mathbf{x}_1$ to the original state $\mathbf{x}_0$.
Similar to the denoising matching term, we define the loss $\mathcal{L}_1$ as the negative of reconstruction term of ELBO as Equation \eqref{eq: L_1},
\begin{equation} \label{eq: L_1}
\small
\begin{aligned}
    \mathcal{L}_1 &\triangleq -\mathbb{E}_{q(\mathbf{x}_{1}|\mathbf{x}_0)}\left[\log p_{\theta}(\mathbf{x}_0|\mathbf{x}_1)\right]\\
    &\propto -\mathbb{E}_{q(\mathbf{x}_{1}|\mathbf{x}_0)}\left[-\dfrac{1}{2\sigma_t^2} \parallel \boldsymbol{\mu}_\theta (\mathbf{x}_1, 1) - \mathbf{x}_0 \parallel_2^2 \right] \\    
    &\Leftrightarrow \mathbb{E}_{q(\mathbf{x}_{1}|\mathbf{x}_0)}\left[\parallel \hat{\mathbf{x}}^{obs}_\theta(\mathbf{x}_1, 1) - \mathbf{x}^{obs}_0 \parallel_2^2  \right] \\
    &\Rightarrow \mathbb{E}_{q(\mathbf{x}_{1}|\mathbf{x}_0)}\left[\parallel M^c\odot\hat{\mathbf{x}}_\theta(\mathbf{x}_1, 1)-M^c\odot\mathbf{x}_0 \parallel_2^2 \right],
\end{aligned}
\end{equation}
where we estimate the log-likelihood of $\log p_{\theta}(\mathbf{x}_0|\mathbf{x}_1)$ thanks to the probability density function of Gaussian distribution
$p_\theta(\mathbf{x}_0|\mathbf{x}_1)=\mathcal{N}(\mathbf{x}_0; \boldsymbol{\mu}_\theta (\mathbf{x}_1, 1), \sigma_t^2 \mathbf{I})$.
We denote the output of reconstruction network as $\hat{\mathbf{x}}_\theta(\mathbf{x}_1, 1)=\boldsymbol{\mu}_\theta (\mathbf{x}_1, 1)$. 
We also only constrain the generation of observed slots. 

\subsubsection{Overall diffusion imputation objective}
The final objective for diffusion imputation optimization is as simple as Equation \eqref{eq: diffusion_loss_ELBO}, that minimizes the L2 distance between available modalities and the output of reconstruction network $\hat{\mathbf{x}}_\theta(\mathbf{x}_t, t)$ at each step $t$. 
\begin{equation} \label{eq: diffusion_loss_ELBO}
\small
\begin{aligned}
   & \arg \max_\theta \text{  Masked-ELBO} 
    \Leftrightarrow  \arg \max_\theta \left[ -\mathcal{L}_1-\Sigma_{t=2}^T \mathcal{L}_t \right] \\
    & \Leftrightarrow  \arg \max_\theta \left[- \Sigma_{t=1}^T \mathcal{L}_t \right] 
    \Leftrightarrow  \arg \min_\theta \left[ \Sigma_{t=1}^T \mathcal{L}_t \right]
\end{aligned}
\end{equation}

\subsection{Privacy Preservation Discussion} \label{supp:privacy discussion}
\subsubsection{Structural privacy}
Structural privacy includes relations and triples. 
Since relation IDs and relation embeddings remain in their own client and do not participate in communication, one client cannot be aware of that of other clients.
As for triples, since the relation types of clients have no overlapping and the relation embeddings are not shared, the relation between two entities belonging to its owner cannot be inferred from other clients.

\subsubsection{Visual and textual privacy}
We preserve the $E_m^c$ from open-sourced pre-trained models in their own client, thus the visual and textual features cannot be accessed by other clients.
In the imputation process, the parameters of the diffusion models are not shared, for privacy-preserving of local knowledge and reducing communication costs as well.
Moreover, our missing modality imputation objective $\mathcal{L}_{DI}$ aims to maintain the semantic consistency of the available ones instead of bringing extra semantics from other clients, which also preserves semantic privacy. 

After training, the structural entity embeddings $\mathbf{S}^c$ of clients tend to be similar, while the sensitive visual and textual semantics in $\mathbf{V}^c$ and $\mathbf{D}^c$ remain local and have not been transmitted.
In future work, provable privacy-preserving techniques such as differential privacy that add noise to the uploaded gradients can be applied in MMFeD3-HidE due to our robustness to noise in diffusion model and heterogeneous clients in federated optimization.

\section{Experiments} \label{supp:experiments}
\input{tables/full_fusion_result}
\input{figs/denoise_zhutu}

\input{figs/paras5}
\subsection{Baselines construction} \label{supp:baselines}
We experiment with baselines from three groups:

\textit{(1) MKGC baselines:}
We include various multimodal fusion functions $\Phi(\cdot)$ based on existing MKGC methods. 
\begin{itemize}
    \item \textbf{Average}: entity embeddings are average of three embeddings as $\mathbf{E}^c=w_s\mathbf{S}^c+w_v\mathbf{V}^c+w_d\mathbf{D}$, of which the weights $w_m$ are $1/3$.
    \item \textbf{Weighted}: entity embeddings are weighted average of three embeddings, where $w_m$ in the average fusion above are trainable parameters.
    \item \textbf{Concatenation}: entity embeddings are the mapped projection of three concatenated embeddings as $\mathbf{E}^c=\text{MLP}([\mathbf{S}^c||\mathbf{V}^c||\mathbf{D}^c])$.
    \item \textbf{Split}: modality split paradigm does not fuse multimodal embeddings and takes an average of three prediction probabilities from each modality as the final prediction, as $p^c(t|(h,r)_i)=\sum_m\text{softmax}(f^c(h,r,e;\mathbf{E}^c_m,\mathbf{R}^c_m)/3$, where $\mathbf{E}^c_m \in \{\mathbf{S}^c,{\mathbf{V}}^c,{\mathbf{D}}^c\}$.
    \item \textbf{Gated}: gated fusion exploits structural embeddings to filter visual and textual information as $\mathbf{V}^c=\mathbf{W}_0(\sigma(\mathbf{W}_1\mathbf{S}^c) * \mathbf{W}_2\mathbf{V}^c)$ (visual modality as example), then utilize the average fusion of three embeddings as entity embedding. 
\end{itemize}

\textit{(2) Federated learning baselines:} 
We augment the MMFedE backbone with FedKGC or FL baselines by adding their additional objectives to MMFedE. 
\begin{itemize}
    \item \textbf{MMFedE}: optimize the federated MKGs with $\mathcal{L}_{KGC}^c$.
    \item \textbf{MMFedEC}: optimize with extra contrastive objective of reconstructed entity embeddings $\mathcal{L}^c_{EC}=-\log\frac{\exp(s(\hat{\mathbf{E}}_{ro}^c,\mathbf{E}_{ro}^s)/\tau}{\exp(s(\hat{\mathbf{E}}_{ro}^c,\mathbf{E}_{ro}^s)/\tau)+\exp(s(\hat{\mathbf{E}}_{ro}^c,\hat{\mathbf{E}}_{ro-1}^c)/\tau)}$, where $\hat{\mathbf{E}}^c$ are the imputed local entity embeddings, $\mathbf{E}^s$ are the incomplete global ones, and $ro$ is the current round.
    \item \textbf{MMFedProx}: optimize with extra proximal term $\mathcal{L}^c_{prox}=||\mathbf{E}^s-\hat{\mathbf{E}}^c||^2_2$.
    \item \textbf{MMFedLU}: optimize with extra global KGC objective $\mathcal{L}_{KGC}^{s,c}$ and logit distillation objective $\mathcal{L}_{LU}^c = \sum_{(h,r,t)_i} D_\text{KL}(p^c(t|(h,r)_i), p^{s,c}(t|(h,r)_i))$.
\end{itemize}

\textit{(3) Incomplete Multimodal Learning (IMML) baselines:}
We denote the baselines as $\theta$ in $\hat{\mathbf{H}}^c=\theta(\mathbf{H}^c)$, and optimize them with masked objective $\mathcal{L}_{recon}^c=\parallel M^c\odot\hat{\mathbf{H}}^c - M^c\odot\mathbf{H}^c \parallel_2^2$
\begin{itemize}
    \item \textbf{AE}: AE has 3-layer MLPs activated by ReLU and each layer with input and output size [512, 256,128].
    \item \textbf{CRA}: CRA is 3 cascaded blocks of AEs with residual connection, where the AEs are as above.
    \item \textbf{MMIN}: MMIN has a CRA encoder and a CRA decoder, and the output of the encoder and decoder are constrained by cycle-consistency imputation loss in two directions as $\parallel M^c\odot\hat{\mathbf{H}}^{c}_f - M^c\odot\mathbf{H}^c \parallel_2^2+\parallel M^c\odot\hat{\mathbf{H}}^{c}_b - M^c\odot\mathbf{H}^c \parallel_2^2$, where $\hat{\mathbf{H}}^{c}_f$ and $\hat{\mathbf{H}}^{c}_b$ are the output of encoder and decoder respectively.
\end{itemize}

\subsection{FedMKGC with full modalities} \label{supp:full modalities}
Table \ref{tab:fusionresults} shows the FedMKGC performance with full modalities of MKGC baselines in independent learning, federated learning, and centralized learning, respectively.

\textbf{Federated Setting: }{\textit{Compared to S-Ind}} (the independently trained structural model), MMInd with different MKGC baselines generally performs better, demonstrating the benefits of multimodal information in complementing structural information.
{\textit{Comparing MMInd, MMFedE, and MMCen}}, the centralized MMCen, which trains a unified global model on all triples, images, and descriptions, performs the best.
MMFedE outperforms MMInd, showing the effectiveness of our aggregation strategy for jointly training client MKGs.
However, MMFedE does not surpass MMCen, likely because its basic aggregation is insufficient to handle the heterogeneous partitioned multimodal information against the unified non-partitioned one in MMCen.

\textbf{MKGC baselines: }
Different MKGC baselines perform diversely across settings. Average and split fusion achieve the best metrics most often in the MMInd group, and weighted fusion does so in the MMFed and MMCen groups.
This suggests the fusion strategies differ in adaptive ability across clients, while weighted fusion adapts best and stays the most stable. Thus, we adopt weighted fusion as the default in the following experiments.
Moreover, Gated fusion performs worse in MMFedE than in MMInd and MMCen, likely because it has more parameters than other fusion paradigms, and the transmission of projection weights in MMFedE is insufficient for better global performance.

\subsection{Effect of Diffusion Reconstruction Network} \label{supp: diff recon net}
We replace the reconstruction network $\theta$ in the reverse process of our diffusion imputation with various reconstruction networks, including AE, MLP, and Multi-head Attention (MHA).
As shown in Figure \ref{fig:diff_recon}, CRA shows the best performance in three datasets, while MHA shows the second-best performance.
The CRA has shown better incomplete multimodal learning ability for iteratively modeling the residual between the incomplete multimodal and the reconstructed one.
The MHA also has strong potential as the denoising network in the diffusion imputation model, probably because of better cross-modal and intra-modal interaction.

\subsection{Hyperparameter Sensitivity Analysis} \label{supp: hyper-paras}
\revision{
We conduct a hyperparameter sensitivity analysis by varying key parameters and reporting their impact on MRR, as shown in Figure \ref{fig:paras5}. All hyperparameters were determined based on performance on the validation set.}

\revision{
Regarding the diffusion module, the model exhibits strong robustness to variations in the imputation weight $\lambda$ and step counts. Specifically, the performance curve for $\lambda$ is bell-shaped with a peak at $2^0$, suggesting that a balanced contribution from the imputation objective best aids the primary KGC task without dominating the gradient. Similarly, the method remains stable across different inference and diffusion steps, achieving optimal results with a small number of steps (e.g., 5 inference steps), confirming HidE's efficiency in manifold mapping.
Regarding the federated distillation, the loss ratios $\eta$ and $\mu$ require careful tuning to balance global knowledge transfer and local personalization. As observed, the feature distillation weight $\eta$ peaks at a smaller value ($10^{-4}$), indicating that subtle feature alignment is sufficient, while excessive constraints (larger $\eta$) may hinder client-specific representation learning. In contrast, logit distillation favors a larger weight ($\mu=2$), but performance drops sharply when $\mu \ge 4$, implying that over-regularization by the server's logits can negatively impact local model convergence.}

\input{figs/case}


\subsection{Case Study}
\revision{
To intuitively illustrate our method's efficacy, we present a case study in Figure \ref{fig:case}. 
The structural, visual, and textual knowledge of entities is distributed across clients, which limits the reasoning capability of isolated local models due to fragmented information.
During the test phase, although specific modalities remain inherently missing, MMFeD3-HidE leverages hyper-modal diffusion imputation and federated dual-distillation to reconstruct comprehensive semantic embeddings.
For the query predicting the brand associated with \textit{Gianna Bryant}, the local \textit{MKG 2} holds limited knowledge about her father \textit{Kobe Bryant} and his links to \textit{NBA} and \textit{Nike}, thus ranking \textit{Adidas} and \textit{Coke Cola} first.
However, after federated learning and diffusion imputation, MMFeD3-HidE recovers the missing discriminative semantics with a complete hyper-modal embedding, correcting the prediction to rank the ground-truth \textit{Nike} at the top.}

%% file: tables/notations.tex
\begin{table}[!t]
\centering
\small
\caption{Notations used in the paper.}
\label{tab: notations}
\resizebox{\linewidth}{!}{
\begin{tabular}{p{1.3cm}|p{6.3cm}}
\toprule
\centering$\mathcal{C}$ & The set of client MKGs. \\
\centering$\mathcal{G}^c$ & The client multimodal knowledge graph. \\
\centering$\mathcal{E}^c,\mathcal{R}^c,\mathcal{T}^c$ & The set of entities, relations, triples of client. \\

\centering$\mathcal{V}^c,\mathcal{D}^c$ & The set of entity images and descriptions. \\
\centering$M_m^c$ & Modality $m$ availability mask in client $c$. \\

\centering $N_i$ & Negative sample set for the $i$-th triple. \\
\centering$f(\cdot)$ & The score function for triple. \\
\centering $\Phi(\cdot)$ & The multimodal fusion method. \\
\midrule
\centering $\mathbf{S}^s,\mathbf{S}^c$ & Server/Client structural entity embeddings. \\
\centering $\mathbf{E}^c,\hat{\mathbf{E}}^c$ & Client incomplete/Imputed fused entity embeddings. \\
\centering$\mathbf{W}_{m}$ & Server modality $m$ projection weight.\\
\centering$\mathbf{W}_{m}^c$ & Client $c$ modality $m$ projection weight.\\
\centering ${E}^c_m$ & Entity $m$ modal features in client $c$. \\
\centering $\mathbf{V}^c,\mathbf{D}^c$ &Visual/Textual entity embeddings in client $c$. \\
\centering $\mathbf{R}^c$ & Relation embeddings in client $c$. \\

\centering $\mathbf{H}^c,\hat{\mathbf{H}}^c$ & Incomplete/Imputed hyper-modal entity embeddings. \\

\centering $\mathbf{E}^{s,c}$ & Server incomplete fused entity embeddings. \\

\bottomrule
\end{tabular}}

\end{table}

%% file: algs/alg_mmfedcd.tex
\begin{algorithm}[!t]     
\small
    \SetKwInOut{Input}{Input}
    \SetKwInOut{Output}{Output}
    \SetKwInOut{Require}{Require}
    \Input{Client set $\mathcal{C}=\{\mathcal{G}^c\}$, mapping matrix $\mathbf{P}$, existence vector $\mathbf{v}$, communication rounds $R$}
    Server initialize $\mathbf{S}_0^s$, $\mathbf{W}_{v,0}$, $\mathbf{W}_{d,0}$;\\
    \For{$ro$ in $R$}{
    Sample a client subset $\mathcal{C}_{ro}\subseteq\mathcal{C}$;\\
    \For{$\mathcal{G}^c \in \mathcal{C}_{ro}$}{
    Server distributes $\mathbf{S}_{ro}^c=\mathbf{P}^c\mathbf{S}_{ro}^s$, $\mathbf{W}_{v,ro}$, $\mathbf{W}_{d,ro}$;\\
    HidE processes $\hat{\mathbf{E}}^c_{ro}=HidE(\mathbf{E}^c_{ro})$\\
    Calculate $\mathcal{L}_{DI}(\mathbf{E}^c_{ro};\theta)$;\\
    Calculate $\mathcal{L}_{KGC}^c(\hat{\mathbf{E}}^c_{ro},\mathbf{R}^c_{ro})$;\\ 
    Calculate $\mathcal{L}_{KGC}^{s,c}(\mathbf{E}^{s,c}_{ro},\mathbf{R}^c_{ro})$;\\
    Calculate $\mathcal{L}^c_{FD}(\hat{\mathbf{E}}^c_{ro},\mathbf{E}^{s,c}_{ro})$;\\
    Calculate $\mathcal{L}_{LD}^c(\hat{\mathbf{E}}^c_{ro},\mathbf{E}^{s,c}_{ro},\mathbf{R}^c_{ro})$;\\
    Calculate $\mathcal{L}^c=\mathcal{L}^c_{KGC} + \mathcal{L}^{s,c}_{KGC} + \lambda  \mathcal{L}^c_{DI} + \mu \mathcal{L}_{LD}^c + \eta \mathcal{L}_{FD}^c$;\\
    End-to-end Update $\mathbf{S}^c_{ro+1}, \mathbf{W}^{c}_{v,ro+1}, \mathbf{W}^{c}_{d,ro+1},\theta,\mathbf{R}^c$, $\mathbf{S}^{s,c}_{ro+1}, \mathbf{W}^{s,c}_{v,ro+1}, \mathbf{W}^{s,c}_{d,ro+1}\leftarrow \arg \min \mathcal{L}^c$;\\
    Client uploads $\mathbf{S}^{s,c}_{ro+1}, \mathbf{W}^{s,c}_{v,ro+1}, \mathbf{W}^{s,c}_{d,ro+1}$;
    }
    $\mathbf{S}_{ro+1}^s=(\mathds{1}\oslash\sum_{\mathcal{C}_{ro}} \mathbf{v}^c)\otimes \sum_{\mathcal{C}_{ro}} {\mathbf{P}^{c}}^\top \mathbf{S}_{ro+1}^{s,c} $;\\
    \For{$m \in \{v,d\}$}{
        $\mathbf{W}_{m,ro+1} = \sum_{\mathcal{C}_{ro}} \alpha^{c} \mathbf{W}_{m,ro+1}^{s,c}$;
    }
    }
        \caption{MMFeD3-HidE Pipeline}
    \label{alg: pipeline hide-mmfed3}
\end{algorithm}

%% file: tables/full_fusion_result.tex
\begin{table*}[!t]
\centering
\caption{FedMKGC performance with fully available entity descriptions and images. 
The descriptions and images in group MMInd and MMFedE are partitioned between clients.
In group MMCen, the descriptions and images of all client KGs are the same.
}
\label{tab:fusionresults}
\resizebox{\linewidth}{!}{
\setlength{\tabcolsep}{1.7mm}{\begin{tabular}{llcccccccccccccc}

\toprule
\multirow{2}{*}{FL}&\multirow{2}{*}{MKGC} &  \multicolumn{4}{c}{FB15K-237-Fed3} & \makebox[0.01\textwidth][c]{} & \multicolumn{4}{c}{FB15K-237-Fed5} & 
\makebox[0.01\textwidth][c]{} & 
\multicolumn{4}{c}{FB15K-237-Fed10}\\
\cline{3-6}
\cline{8-11}
\cline{13-16}
& & Hits@1 & Hits@3 & Hits@10 & MRR &  & Hits@1 & Hits@3 & Hits@10 & MRR &  & Hits@1 & Hits@3 & Hits@10 & MRR\\

\midrule
S-Ind & - & 0.231 & 0.390 & 0.547 & 0.338 &  & 0.220 & 0.373 & 0.532 & 0.325 &  & 0.205 & 0.361 & 0.517 & 0.310\\
\midrule
\multirow{5}{*}{MMInd}
 & Avg  & 0.232 & \textbf{0.408} & \textbf{0.576} & \textbf{0.349} &  & 0.221 & 0.392 & 0.555 & \textbf{0.335}  &  & \textbf{0.218} & 0.384 & 0.551 & \textbf{0.331} \\
&Weighted  & \textbf{0.233} & 0.405 & 0.572 & 0.348 &  & \textbf{0.222} & 0.391 & 0.551 & 0.334 &  & 0.217 & 0.381 & 0.549 & 0.329 \\
&Concat    & 0.231 & 0.396 & 0.562 & 0.343 &  & 0.210 & 0.368 & 0.535 & 0.319  &  & 0.203 & 0.357 & 0.526 & 0.311 \\
&Split     & 0.222 & 0.413 & \textbf{0.576} & 0.346 &  & 0.213 & \textbf{0.398} & \textbf{0.559} & 0.333  &  & 0.212 & \textbf{0.391} & \textbf{0.560} & \textbf{0.331} \\
&Gated     & 0.230 & 0.403 & 0.569 & 0.346 &  & 0.217 & 0.385 & 0.552 & 0.330  &  & 0.215 & 0.375 & 0.542 & 0.325 \\
\midrule
\multirow{5}{*}{MMFedE}&Avg & \textbf{0.237} & \textbf{0.438} & 0.610 & 0.366 &  & 0.232 & 0.436 & 0.609 & 0.363  &  & 0.217 & 0.428 & 0.606 & 0.352 \\
&Weighted & \textbf{0.237} & 0.437 & \textbf{0.612} & \textbf{0.367} &  & \textbf{0.234} & \textbf{0.439} & \textbf{0.613} & \textbf{0.365}  &  & \textbf{0.221} & \textbf{0.434} & \textbf{0.614} & \textbf{0.357} \\
&Concat   & 0.233 & 0.436 & 0.602 & 0.362 &  & 0.230 & 0.430 & 0.598 & 0.358  &  & 0.219 & 0.422 & 0.595 & 0.350 \\
&Split    & 0.234 & 0.427 & 0.597 & 0.359 &  & 0.227 & 0.425 & 0.597 & 0.354  &  & 0.220 & 0.416 & 0.592 & 0.349 \\
&Gated    & 0.211 & 0.406 & 0.587 & 0.339 &  & 0.192 & 0.394 & 0.577 & 0.324  &  & 0.189 & 0.396 & 0.584 & 0.324 \\
\midrule
\multirow{5}{*}{MMCen}&Avg  & 0.245 & 0.440 & 0.613 & 0.371 &  & {0.252} & 0.449 & \textbf{0.627} & \textbf{0.380}  &  & {0.241} & \textbf{0.448} & \textbf{0.629} & \textbf{0.375} \\
&Weighted  & \textbf{0.246} & \textbf{0.441} & \textbf{0.616} & \textbf{0.373} &  & 0.251 & \textbf{0.451} & \textbf{0.627} & \textbf{0.380}  &  & 0.240 & 0.447 & \textbf{0.629} & 0.373 \\
&Concat    & 0.230 & 0.426 & 0.602 & 0.358 &  & 0.248 & 0.444 & 0.622 & 0.376  &  & 0.234 & 0.441 & 0.625 & 0.368 \\
&Split     & 0.225 & 0.423 & 0.595 & 0.353 &  & 0.237 & 0.442 & 0.611 & 0.368  &  & 0.228 & 0.437 & 0.614 & 0.362 \\
&Gated     & 0.245 & 0.438 & 0.613 & 0.371 &  & \textbf{0.253} & 0.446 & 0.620 & 0.379  &  & \textbf{0.243} & 0.442 & 0.621 & 0.373   \\   
\bottomrule
\end{tabular}}
}

\end{table*}

%% file: figs/denoise_zhutu.tex
\begin{figure*}[t]
     \centering
         \centering
         \includegraphics[width=\textwidth]{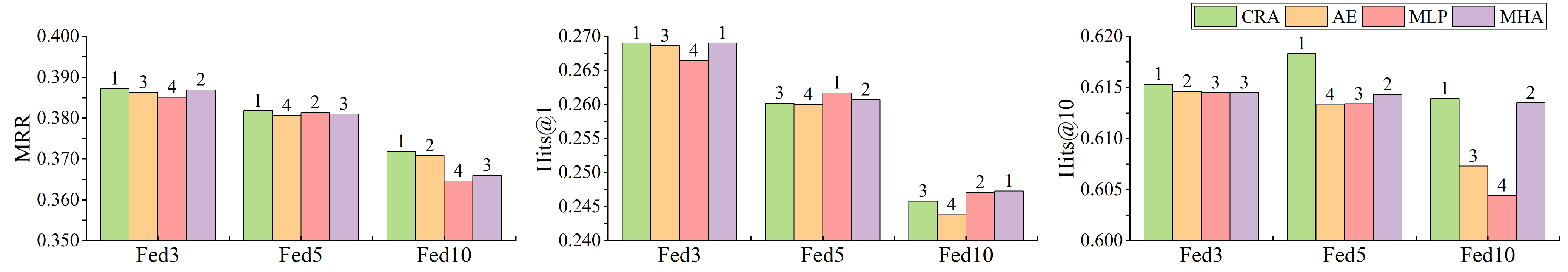}
         \caption{MMFeD3-HidE performance with different reconstruction networks. The rankings of each reconstruction network are labeled above the bars.}
    \vspace*{\fill}
    
\label{fig:diff_recon}
\end{figure*}

%% file: figs/paras5.tex
\begin{figure*}[t]
     \centering
         \centering
         \includegraphics[width=\textwidth]{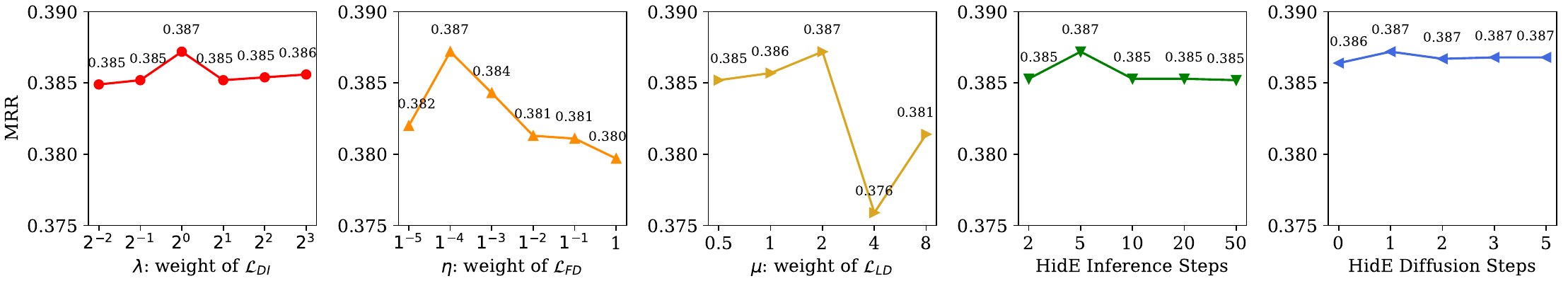}
         \caption{Performance of MMFeD3-HidE with different hyper-parameters.}
    \vspace*{\fill}
\label{fig:paras5}
    \vspace{-15pt}

\end{figure*}

%% file: figs/case.tex
\begin{figure}[t]
     \centering
         \centering
         \includegraphics[width=0.45\textwidth]{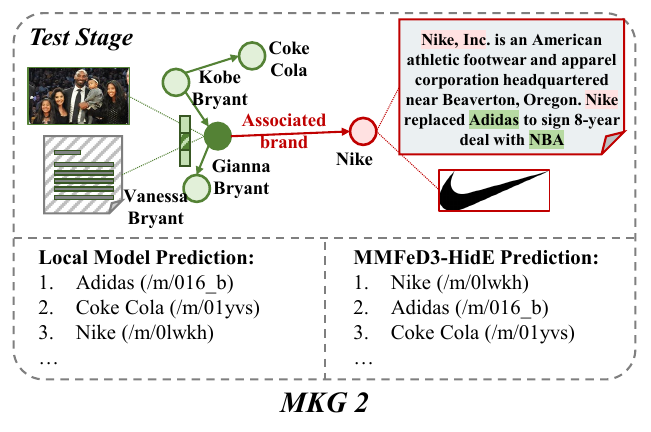}
         \caption{Toy case for test phase of MMFeD3-HidE. }
    \vspace*{\fill}
    \vspace{-15pt}
\label{fig:case}
\end{figure}